\documentclass[journal]{IEEEtran}
\usepackage{graphicx}
\usepackage{color,soul}
\usepackage{multirow,booktabs,makecell,bigstrut}
\usepackage[hidelinks]{hyperref}
\usepackage{balance}

\usepackage{cite}

\ifCLASSINFOpdf
\else
\fi
\usepackage{amsmath, amsfonts, amssymb, bm}
\newcommand{\etal}{\textit{et al. }}
\newcommand{\mathopr}[1]{\mathtt{#1}}

\begin{document}
%
\title{Scale-Aware Dynamic Network for Continuous-Scale Super-Resolution}
%
%
%

\author{Hanlin~Wu,~\IEEEmembership{Student Member,~IEEE},
  Ning~Ni,
  Libao~Zhang,~\IEEEmembership{Member,~IEEE}
  \thanks{
    This work was supported in part by the Beijing Natural Science Foundation under Grant L182029, in part by the National Natural Science Foundation of China under Grant 61571050 and Grant 41771407.
    \textit{(Corresponding author: Libao Zhang.)}

    The authors are with the School of Artificial Intelligence, Beijing Normal University, Beijing 100875, China. (e-mail: libaozhang@bnu.edu.cn).}
}

%
%

\markboth{}%
{}
%



\maketitle



\begin{abstract}
  Single-image super-resolution (SR) with fixed and discrete scale factors has achieved great progress due to the development of deep learning technology. However, the continuous-scale SR, which aims to use a single model to process arbitrary (integer or non-integer) scale factors, is still a challenging task. The existing SR models generally adopt static convolution to extract features, and thus unable to effectively perceive the change of scale factor, resulting in limited generalization performance on multi-scale SR tasks. Moreover, the existing continuous-scale upsampling modules do not make full use of multi-scale features and face problems such as checkerboard artifacts in the SR results and high computational complexity. To address the above problems, we propose a scale-aware dynamic network (SADN) for continuous-scale SR. First, we propose a scale-aware dynamic convolutional (SAD-Conv) layer for the feature learning of multiple SR tasks with various scales. The SAD-Conv layer can adaptively adjust the attention weights of multiple convolution kernels based on the scale factor, which enhances the expressive power of the model with a negligible extra computational cost. Second, we devise a continuous-scale upsampling module (CSUM) with the multi-bilinear local implicit function (MBLIF) for any-scale upsampling. The CSUM constructs multiple feature spaces with gradually increasing scales to approximate the continuous feature representation of an image, and then the MBLIF makes full use of multi-scale features to map arbitrary coordinates to RGB values in high-resolution space. We evaluate our SADN using various benchmarks. The experimental results show that the CSUM can replace the previous fixed-scale upsampling layers and obtain a continuous-scale SR network while maintaining performance. Our SADN uses much fewer parameters and outperforms the state-of-the-art SR methods. The code and pretrained models are available at \url{https://github.com/hanlinwu/SADN}.
\end{abstract}

\begin{IEEEkeywords}
  Image restoration, super-resolution, scale-aware, dynamic convolution, continuous-scale, implicit function.
\end{IEEEkeywords}

%
\IEEEpeerreviewmaketitle

\section{Introduction}

Single-image super-resolution (SISR) aims to restore a high-resolution (HR) image by adding high-frequency details to its low-resolution (LR) counterpart. SISR is a long-standing fundamental task of computer vision, and it is also a challenging and ill-posed problem. Limited by the resolution of imaging equipment, SISR has broad application prospects in many fields, such as medical imaging \cite{huang2017simultaneous,mahapatra2019image} and remote sensing imaging \cite{wu2020remote, dong2020remote}. In these application scenarios, it makes sense to enlarge LR images to multiple scales because images with different scales will present different details. Furthermore, these scales should not be restricted to integers but should also include arbitrary positive numbers. However, it is impossible to train an SR model for each scale in practice, which will lead to problems of low computational efficiency and excessive storage space. Therefore, it is of great significance to construct a single model that can super-resolve any scale factor.

\begin{figure}[t]
   \centering
   \includegraphics[width=0.95\linewidth]{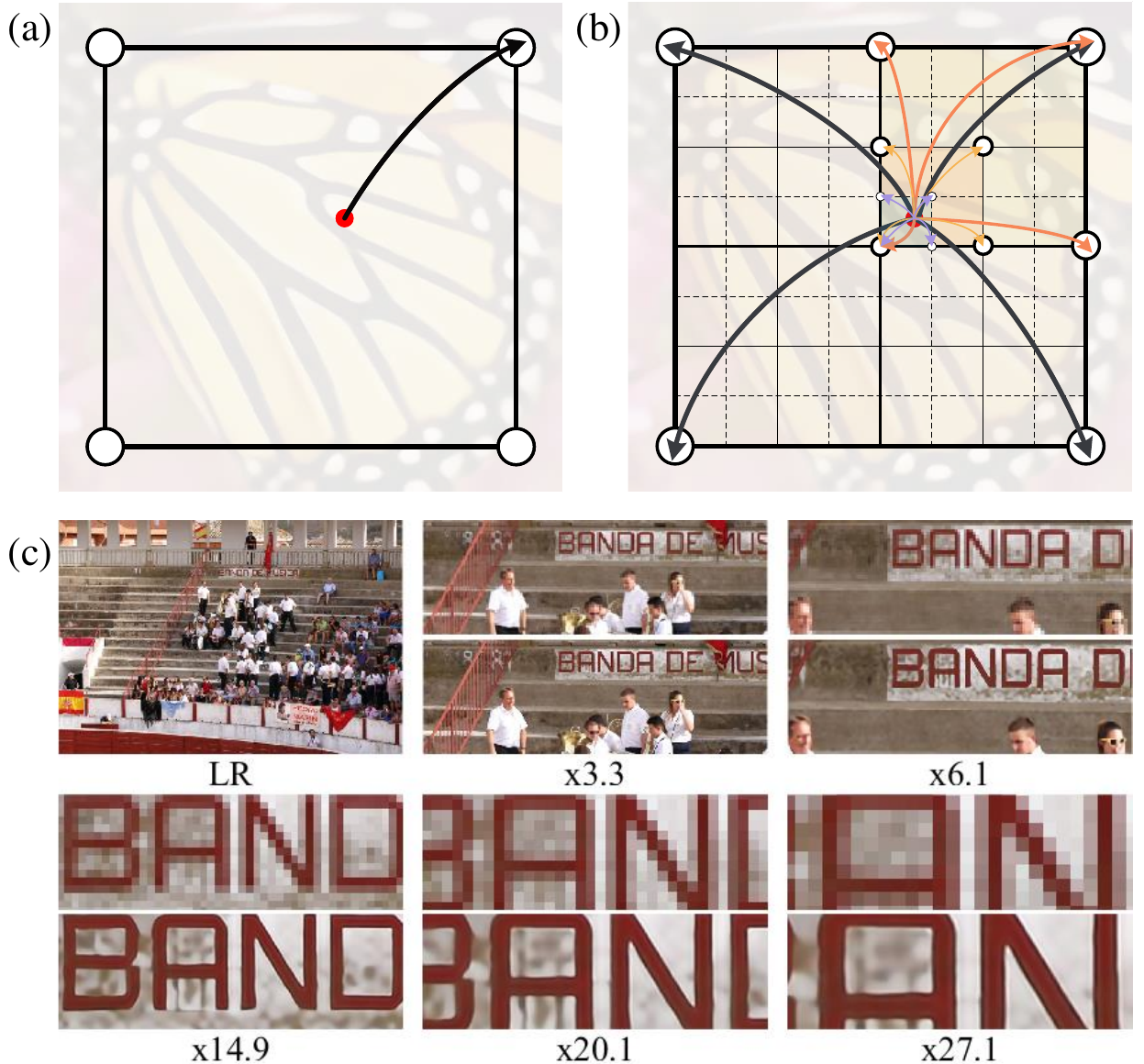}
   \caption{(a) Meta-SR \cite{Hu2019} and the LIIF \cite{chen2021learning} only use one feature vector in a single-scale space. (b) Our proposed MBLIF uses a set of feature vectors from feature maps with multiple scales. (c) Visual results of our continuous-scale SR. The upper image patches are LR inputs, and the lower patches are SR results.}
   \label{fig:intro}
   \vspace{-1em}
\end{figure}

Deep learning (DL) technology has demonstrated its prominent superiority compared to other machine learning algorithms in SISR \cite{jin2019flexible,yan2019deep,li2021learning}. Although deep learning methods greatly improved the performance of SISR, most of them are designed for a single scale factor or multiple fixed scale factors. Early pre-upsampling methods, such as the SRCNN\cite{dong2015image}, use bicubic interpolation at the beginning of the network to achieve arbitrary scale SR. However, the low computational efficiency caused by pre-upsampling makes these models difficult to apply to actual scenes. The post-upsampling methods first extract the features of an LR image and then use the deconvolutional layer \cite{zeiler2010deconvolutional} or sub-pixel layer \cite{shi2016real} at the end of the network to increase the resolution. Sub-pixel layers with different scale factors have different structures and parameters, so the sub-pixel layer can only handle a single scale factor at one time. The deconvolutional layer usually needs to adjust the kernel size according to the scale factor and can only handle integer scale factors. Therefore, a network using a sub-pixel layer or a deconvolutional layer can only be trained with a single integer scale factor, which makes the model unable to be generalized to other scale factors.

At present, only a few studies have sought to use a single model to simultaneously process multiple scale factors. MDSR \cite{Lim2017} uses multiple upsampling branches to process multiple scale factors, but it can only super-resolve trained integer scales. Meta-SR \cite{Hu2019}, which uses a fully connected network to predict the weights of filters for the upsampling of each scale, is a pioneering work for continuous-scale SR. However, Meta-SR only uses the feature vector of a single position (called latent code) in the LR space for prediction, resulting in the latent codes of two adjacent pixels used for prediction perhaps being different. Therefore, discontinuous patterns would appear where the latent code switches, and the generated HR image may have checkerboard artifacts. Chen \etal \cite{chen2021learning} proposed a local implicit image function (LIIF) to generate the continuous representation of an image and used a local ensemble strategy to avoid the artifact problem of Meta-SR. However, the local ensemble strategy brings additional computational costs due to repeated prediction steps. Moreover, both Meta-SR and the LIIF only focus on building an upsampling module but ignore the different requirements for feature extraction in SR tasks for different scales.

To address the above problems, in this paper, we propose a scale-aware dynamic network (SADN) for continuous-scale SR. The SADN consists of two main parts: a scale-aware feature learning (SAFL) part and a continuous-scale upsampling module (CSUM).

It is observed that the convolution kernels of the models trained for different scale factors have similar patterns but have different statistical information \cite{he2019modulating}. Therefore, in the SAFL part, we propose to dynamically predict the convolutional weights used for each scale factor. To this end, inspired by \cite{chen2020dynamic}, we design a scale-aware dynamic convolutional (SAD-Conv) layer that learns to predict deep convolution kernels conditioned on a scale factor. SAD-Conv can increase the network capability by perceiving the input feature and scale factor with a negligible extra computational cost. In addition, we utilize a feedback mechanism \cite{zamir2017feedback, Li2019c} in the feature learning part so that high-level features can recurrently refine low-level features. Such a recurrent structure can significantly deepen the network without increasing the number of parameters.

In the CSUM, inspired by \cite{chen2021learning}, we construct multi-bilinear local implicit functions (MBLIFs) to obtain a continuous representation of an image. Instead of using only one latent code in a single feature space to predict the signal, we propose building multiple feature spaces with different scales to obtain more reliable feature expressions and using a local set of latent codes in each space to construct implicit functions. Fig.\,\ref{fig:intro} demonstrates the differences between the proposed MBLIF, Meta-SR \cite{Hu2019}, and LIIF \cite{chen2021learning}. First, we map the LR features obtained by a feature extractor to multiple HR feature spaces with different scales, such as $\times 1$, $\times 2$, $\times 4$, and $\times 8$. Then, in each feature space, we obtain a continuous feature representation through local bilinear implicit functions. Finally, the outputs of the local bilinear implicit functions in each space are combined using a scale-aware nonlinear function to predict the RGB value. MBLIFs can map any 2D coordinate to an RGB value and effectively avoid the problem of discontinuous patterns. Besides, the CSUM can be easily plugged into other state-of-the-art (SOTA) SR networks. All we need to do is replace the original upsampling module with our proposed CSUM, and we obtain a continuous-scale SR network.

Our main contributions are as follows:

1)  We propose a scale-aware feature extractor incorporating the SAD-Conv. The SAD-Conv can adaptively adjust the filter weights according to the input scale factor, which allows the feature learning part to better adapt to the multi-task learning of various scale factors.

2) We propose an MBLIF, which makes full use of multi-scale features to achieve end-to-end upsampling of arbitrary scale factors. In this way, we can obtain a continuous representation of an image, thereby avoiding the problem of checkerboard artifacts. Furthermore, the proposed CSUM is highly efficient and can be easily embedded in most SR networks.

3) We propose a novel continuous-scale SR framework that is lightweight and efficient. The SADN even outperforms the SOTA fixed-scale SR methods in terms of objective metrics and visual effects.

\section{Related Work}
In this section, we briefly review DL-based SISR methods. More comprehensive reviews can be found in the survey papers \cite{yang2019deep, wang2021deep}.
\subsection{Fixed-Scale SR}

\begin{figure*}
    \centering
    \includegraphics[width=0.96\textwidth]{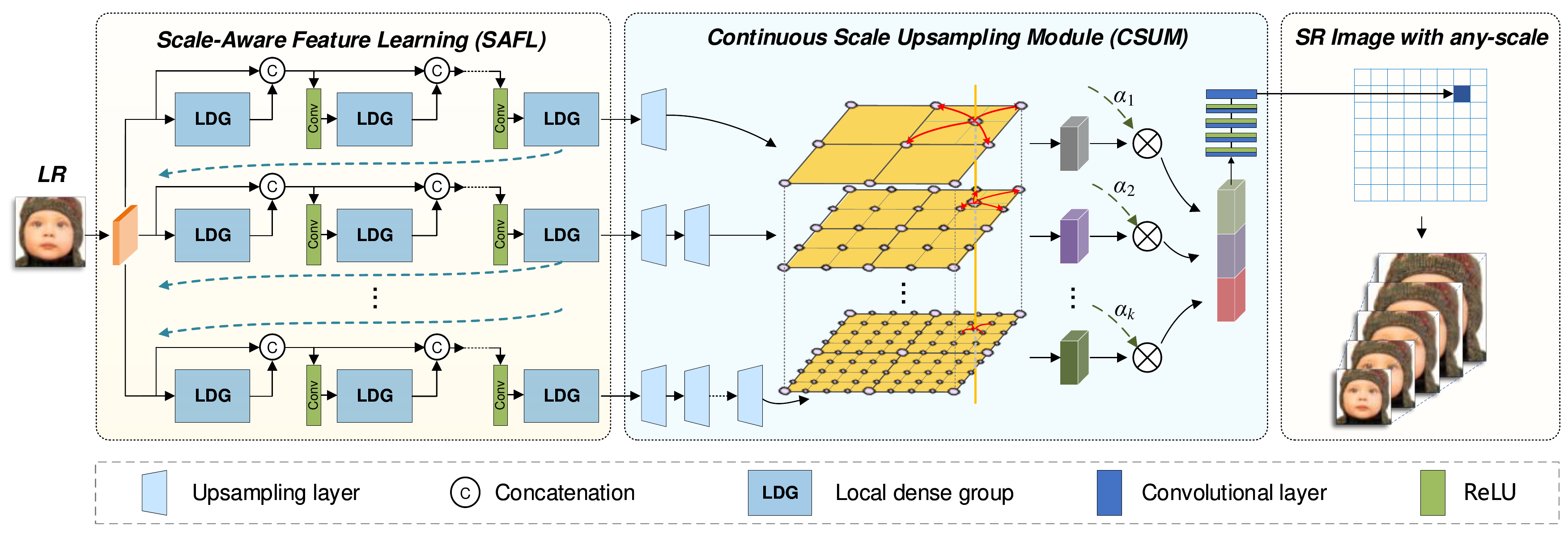}
    \caption{Flowchart of the proposed SADN.}
    \label{fig:flowchart}
    \vspace{-0.5em}
\end{figure*}

Deep neural networks, which have made great progress in SISR, can learn the mapping from LR space to HR space in an end-to-end manner. DL-based methods can be roughly divided into two categories: pre-upsampling models and post-upsampling models.

The pre-upsampling models, such as the SRCNN \cite{dong2015image} and VDSR \cite{kim2016accurate}, first upsample LR images to the desired size and then use deep convolutional networks to restore high-frequency details. However, pre-upsampling causes more operations to be performed in a high-dimensional space, which makes these algorithms computationally expensive and time-consuming.

To improve the computational efficiency, researchers proposed post-upsampling models \cite{dong2016accelerating, ledig2017photo, Zhang2018d, zhang2020accurate} that first learn the features of LR images in a low-dimensional space and then integrate a learnable upsampling layer at the end of the network to complete the SR reconstruction. Post-upsampling has become the mainstream framework in the SR field due to its advantages in training and inference speed. Shi \etal \cite{shi2016real} proposed a learning-based upsampling layer called sub-pixel layer, which generates multiple channels through convolution and then reshapes them to perform upsampling. The deconvolutional layer \cite{zeiler2014visualizing} is also an end-to-end learnable upsampling layer. It increases the resolution by inserting zero values into an LR image and then performs a normal convolution. Zhang \etal \cite{Zhang2018d} proposed a residual dense network (RDN), which combines the advantages of residual blocks \cite{huang2016deep} and densely connected blocks \cite{tong2017image} and further improves the SR performance. Li \etal \cite{Li2019c} introduced a feedback mechanism into the SR task and proposed a lightweight super-resolution feedback network (SRFBN). Soh \etal \cite{soh2020meta} proposed a novel training scheme based on meta-transfer learning that can exploit both external information from a large-scale dataset and internal information from a specific image.

These networks only focus on specific scale factors and treat each scale as an independent task. Retraining a model for each scale factor is extremely time-consuming and consumes too much storage space. Besides, the scale factor should not be limited to an integer but should be any positive real number.

\subsection{Multi-Scale SR}

To the best of our knowledge, VDSR \cite{kim2016accurate} is the first CNN-based model trained on three different scale factors. Lim \etal \cite{Lim2017} proposed the multi-scale SR (MDSR) architecture, which has multiple upsampling branches for different scale factors. Wang \etal \cite{wang2018resolution} proposed a resolution-aware network (RAN) for simultaneously performing SR at multiple scales. Given a scale factor, the RAN divides the SR process into a cascade of intermediate steps. Nevertheless, these structures cannot be applied to continuous-scale SR because they can only handle predefined and trained scale factors.

Subsequently, Hu \etal \cite{Hu2019} proposed a meta-upscaling module to construct a scale-arbitrary super-resolution network (Meta-SR). Meta-SR constructs a coordinate mapping from the LR space to the HR space and then uses a fully connected network to dynamically predict the weights of the filters for the upsampling of each scale. Meta-SR is trained and tested using $30$ scales from $1.1$ to $4.0$ with a stride of $0.1$. However, Meta-SR cannot obtain satisfactory results when handling untrained large-scale factors. The main reason is that Meta-SR only used a single feature vector in the LR space to predict the RGB value in the HR space. In other words, a latent code in the LR space is responsible for generating an image patch in the HR space. This discrete feature expression will cause checkerboard artifacts due to the sudden switching of latent codes.

Recently, inspired by the success of implicit neural representation in 3D tasks \cite{saito2019pifu,chen2019learning}, Chen \etal \cite{chen2021learning} introduced implicit functions to represent pixel-based images continuously. The LIIF essentially still only uses one latent code in the LR space, similar to Meta-SR. Although the LIIF can avoid the problem of discontinuous patterns through the local ensemble strategy, it will bring additional computational complexity because different latent codes are used to calculate the results multiple times. Besides, Behjati \etal \cite{behjati2021overnet} proposed an overscaling module (OSM) that generates overscaled feature maps for continuous-scale upsampling. However, the OSM can only perform the SR of scale factors below a certain maximum value.

While impressive progress has been made, the above continuous-scale SR models use the same feature extractor for all scales, ignoring the needs of different tasks. In addition, the multi-scale features are not fully utilized in the upsampling module.

\section{Methodology}

The architecture of our SADN is shown in Fig.\,\ref{fig:flowchart}. Similar to most DL-based SR methods, the proposed SADN consists of three parts: the head part for shallow feature extraction, the body part for scale-aware feature learning, and the tail part for continuous-scale upsampling.

Let $I_{\mathrm{LR}}$ and $I_{\mathrm{HR}}$ denote the input LR image and the corresponding HR image, respectively. The task of SISR is to generate an image $I_{\mathrm{SR}}$ with the same spatial resolution as $I_{\mathrm{HR}}$. We first extract the shallow features $F_0$ of $I_{\mathrm{LR}}$ as
\begin{equation}
   F_0 = \mathcal{H}(I_{\mathrm{LR}}),
\end{equation}
where $\mathcal{H}$ denotes the operation of the head part in which we use a $3\times 3$ convolutional layer to extract shallow features. The SAFL part takes the scale factor and shallow features as input, contains $T$ iterations, and outputs a feature list with $T$ different levels. Let $r$ denote the scale factor of the current SR task. Then, the body part can be formulated as:
\begin{equation}
   [F_1, F_2,\cdots, F_T] = \mathcal{B}(r, F_0).
\end{equation}
At the end of the network, the extracted multi-level features are passed to the CSUM to generate the super-resolved image $I_{SR}^{(r)}$ with scale factor $r$:
\begin{equation}
   I_{SR}^{(r)} = \mathcal{T}(r, F_1, F_2, \cdots, F_T),
\end{equation}
where $\mathcal T$ denotes the operation of the tail part, which also depends on the scale factor. 

The network is optimized by the $L_1$ loss function. To make the network generalizable across scales, we provide multi-scale data pairs for training to provide sufficient supervision. Let $U(a, b)$ denote a uniform distribution on $[a, b]$, where $a$ and $b$ are the upper and lower bounds of the scale factor used for training, respectively. Then, the loss function is defined as:
\begin{equation}
   \mathcal L(\Theta) = \mathbf E_{r\sim U(a, b), I_{\mathrm{HR}}} \|I_{\mathrm{SR}}^{(r)} - I_{\mathrm{HR}}\|_1,
\end{equation}
where $\Theta$ denotes the network parameter.

\begin{figure}[t]
   \centering
   \includegraphics[width=0.95\linewidth]{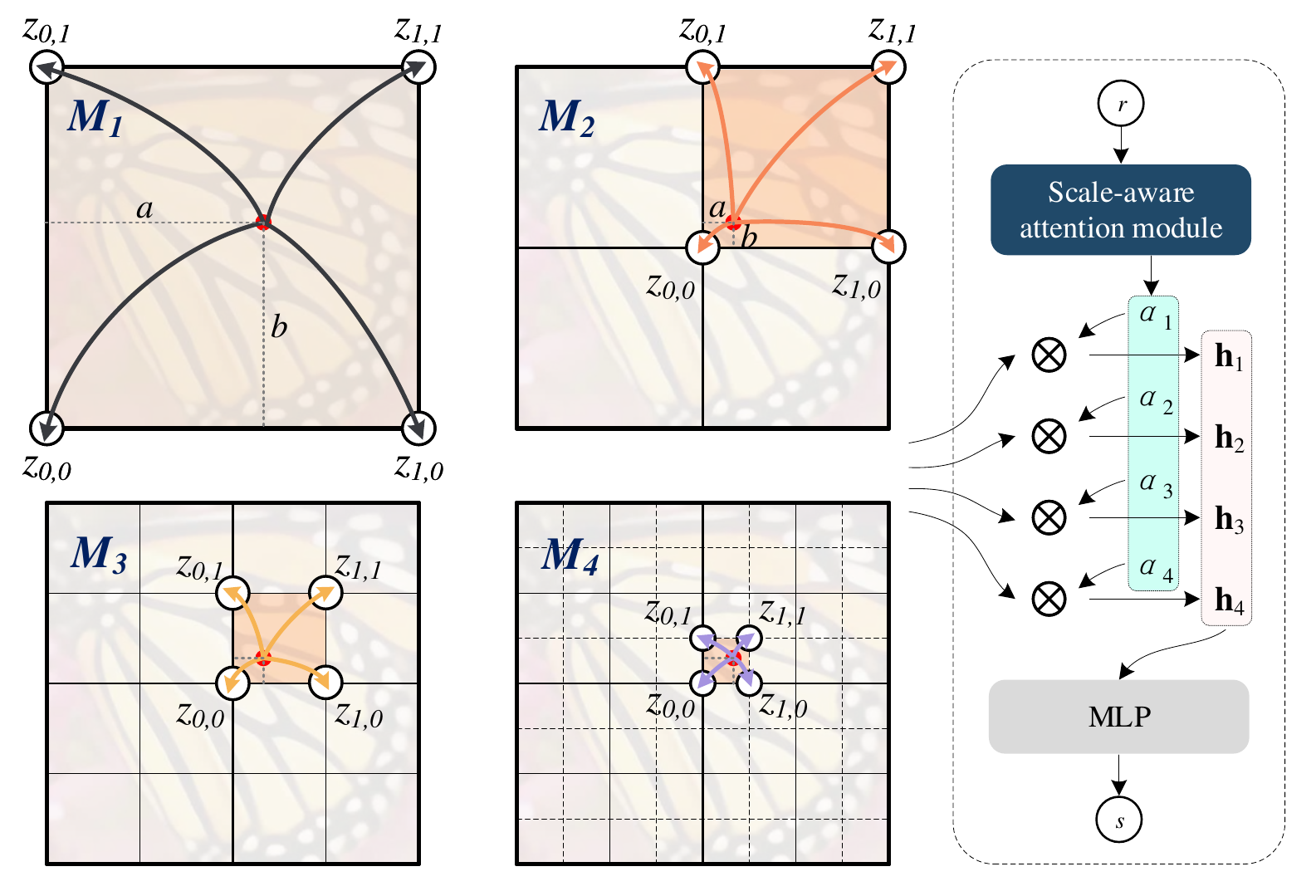}
   \caption{An illustration of the multi-bilinear local implicit function (MBLIF).}
   \vspace{-0.5em}
\end{figure}

\subsection{Multi-Bilinear Local Implicit Function (MBLIF)}

The implicit function is a continuous and differentiable function parameterized by a neural network, which maps a two-dimensional (2D) coordinate to a signal. HR images can be regarded as images with denser sampling points than LR images. Therefore, as long as the continuous representation of an image is obtained, continuous scale upsampling can be achieved. To this end, we propose a novel implicit function called the MBLIF, which is a nonlinear combination of bilinear functions in multi-scale spaces, to learn the continuous representation of an image. Below, we introduce the details of our proposed MBLIF.

Considering that the continuous representation of an image can be viewed as a discrete 2D image with infinitely dense pixels, we use a family of discrete 2D feature maps with increasing resolution to construct the MBLIF. Suppose we have obtained $T$ feature maps $\{M_t\}_{t=1}^T$ with gradually increasing scales $r_1 < r_2 < \cdots < r_T$. Let $f_\theta$ denote the implicit function, and it takes the form:
\begin{equation}
   s = f_{\theta}(x\,;M_1, M_2, \cdots, M_T),
\end{equation}
where $x$ and $s$ denote the 2D coordinate and predicted RGB value, respectively. 
We call the feature vectors in $M_t$ latent codes. Given $\{M_t\}_{t=1}^T$, the function $f_\theta(\,\cdot\,;M_1, M_2, \cdots, M_T)$ can be viewed as a continuous representation of an image. 

We further decompose $f_\theta$ into a nonlinear combination of multiple local implicit feature functions:
\begin{equation}
   s = g_\theta\big(r, h(x, \mathcal Z_1), h(x, \mathcal Z_2), \cdots, h(x, \mathcal Z_T) \big),
\end{equation}
where $h(\,\cdot\,, \mathcal Z_t)$ is the local implicit feature function, and $\mathcal Z_k$ is a set of the four nearest latent codes to $x$ in $M_k$. $g_\theta$ is a scale-aware feature fusion block and takes the scale factor $r$ of the SR task as part of the input. Let $z_{0,0}, z_{0,1}, z_{1,0}, z_{1,1}$ (we omit $t$ and $x$ for clarity) denote the lower left, upper left, lower right, and upper right latent codes of $x$ in $M_t$, respectively. Then, $\mathcal Z_t = \{z_{0,0}, z_{0,1}, z_{1,0}, z_{1,1}\}$. We assume that the feature space of each scale is locally bilinear, so the local bilinear implicit feature function takes the form:
\begin{equation}
   h(x, \mathcal Z_t) = 
   \begin{bmatrix}
      1-a & a
   \end{bmatrix}
   \begin{bmatrix}
         z_{0,0} & z_{0,1} \\
         z_{1,0} & z_{1,1}
   \end{bmatrix}
   \begin{bmatrix}
      1-b\\
      b
   \end{bmatrix},
\end{equation}
where $(a, b)$ denotes the coordinate of $x$ relative to $z_{0,0}$. Note that the relative coordinates are normalized to $[0, 1]$ to eliminate the influence of scale factors.

For the scale-aware feature fusion block $g_\theta$, we use a small multi-layer perceptron (MLP) with two hidden layers to calculate the scale-related attention weights. Then, a five-layer MLP is used to fuse the multi-scale features and produce the final RGB value. Formally, let $\bm\alpha = [\alpha_1, \cdots, \alpha_T]$ be the scale-related attention weights, and it can be calculated as:
\begin{equation}
   \bm\alpha = [\alpha_1, \cdots, \alpha_T] = \mathcal A(1/r),
\end{equation}
where $\mathcal A$ denotes the scale-related attention block. Then, the multi-scale features are corrected using attention weights, $\mathbf h_t = \alpha_t\cdot h(x, \mathcal Z_t)$. Finally, the predicted signal $s$ is computed as:
\begin{equation}
   s = \mathcal F(\mathopr{Concat}(\mathbf h_1, \mathbf h_2, \cdots, \mathbf h_T)),
\end{equation}
where $\mathopr{Concat}$ denotes the concatenation operation and $\mathcal F$ denotes the operation of the five-layer MLP. 

When the query coordinate $x$ moves, the nearest latent code set may change suddenly. However, thanks to the bilinear function, the local implicit feature function changes continuously with respect to the coordinate. Therefore, the MBLIF is a continuous representation of an image.

\subsection{Continuous-Scale Upsampling Module (CSUM)}
Our proposed CSUM accepts $T$ multi-level feature maps with the same size as the LR image as inputs. To obtain multi-scale feature maps $\{M_t\}_{t=1}^T$, we use $T$ parallel upsampling branches with scales $\{r_1,r_2,\cdots, r_T\}$. The output the $t^{\mathrm{th}}$ branch is computed as:
\begin{equation}
   M_t = \mathopr{UP}_t(F_t),
\end{equation}
where $\mathopr{UP}_k$ denotes the operation of the $t^{\mathrm{th}}$ upsampling branch. $M_t\in\mathbb R^{r_t H\times r_t W\times C}$, where $H$ and $W$ denote the height and weight of the LR image, respectively. To reduce the increased number of parameters caused by multiple upsampling branches, we decompose the upsampling operation into multiple steps with small scaling factors to achieve parameter sharing. The operation of the $t^{\mathrm{th}}$ branch can be decomposed as 
\begin{equation}
   \mathopr{UP}_t = \mathopr{SubPixel}_t\circ\mathopr{UP}_{t-1},
\end{equation}
where $\mathopr{SubPixel}_{t}$ denotes a sub-pixel \cite{shi2016real} layer, and it is specific to the $t^{\mathrm{th}}$ upsampling branch. Then, we use the MBLIF to obtain the upsampling results.  
 

Since most existing SR methods only output one feature map in the deep feature extraction part, to make the proposed CSUM be easily plugged into these models, we also present a more general version of the CSUM.  As a special case of the previous CSUM, all elements of the input list can be the same, i.e., $F_1 = F_2 =\cdots =F_T = F$. In this way, the CSUM can be regarded as accepting only one feature map as input. 

\subsection{Scale-Aware Dynamic Convolution (SAD-Conv)}

\begin{figure}[t]
   \centering
   \includegraphics[width=0.95\linewidth]{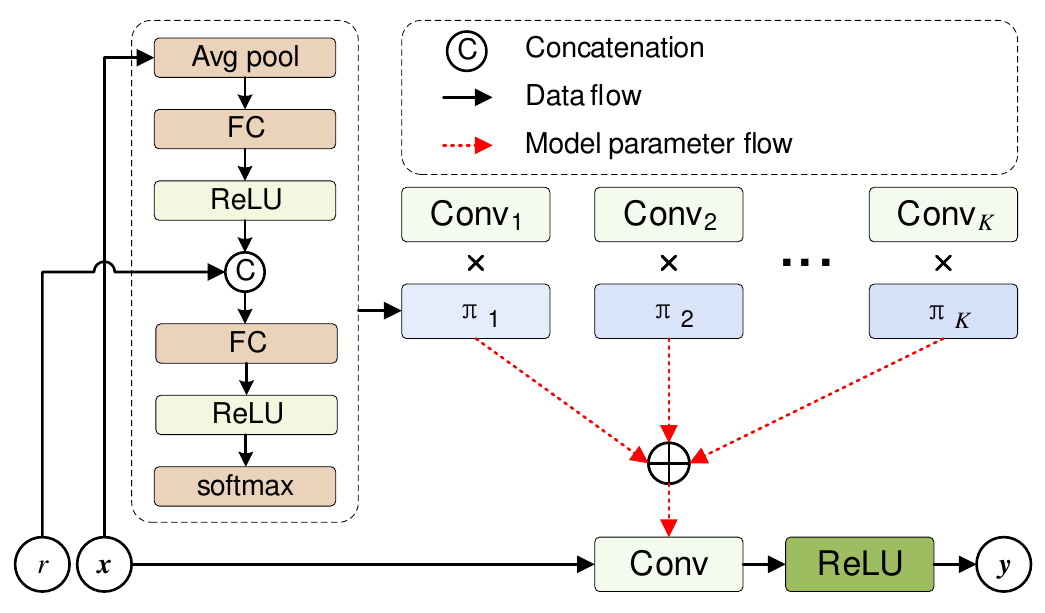}
   \caption{An illustration of the scale-aware dynamic convolution (SAD-Conv).}
   \label{fig:SAD_conv}
   \vspace{-0.5em}
\end{figure}

Although the SR tasks of different scale factors are similar, they also have essential differences, especially when the scale factors are quite different. Extracting scale-specific features is one of the keys to achieving scale-arbitrary SR with a single model. To this end, we propose the SAD-Conv layer, which can dynamically predict the kernel of a convolution operation conditioned on a scale factor. The architecture of the SAD-Conv layer is shown in Fig.\,\ref{fig:SAD_conv}. 

The SAD-Conv layer has $K$ convolution kernels (and biases) with the same kernel size. We adopt a scale-aware kernel attention block to adaptively calculate the attention weights of these kernels according to the input scale factor and feature map. Then, these kernels are fused using these attention weights. Specifically, the kernel $\bm W(\bm y, r)$ and bias $\bm b(\bm y, r)$ of the SAD-Conv layer are functions of the input feature $\bm y$ and the scale factor $r$. The SAD-Conv is defined as follows:
\begin{equation}
   \begin{aligned}
      \bm{W}(\bm y, r) &= \sum_{k=1}^K \pi_k(\bm y, r)\bm W_k,\\
      \bm{b}(\bm y, r) &= \sum_{k=1}^K \pi_k(\bm y, r) \bm b_k. \\
      \mathrm{s.t.\ \ } 0\leq \pi_k(\bm y, &r) \leq 1, \sum_{k=1}^K \pi_k(\bm y, r) = 1.
   \end{aligned}
\end{equation}
where $\pi_k$ denotes the kernel attention weight of the $k^{\mathrm{th}}$ kernel $\bm W_k$ and bias $\bm b_k$. 

We take the squeeze-and-excitation block (SE-block) \cite{hu2018squeeze} as the baseline of our scale-aware kernel attention block. The difference is that we concatenate the output of the first fully connected (FC) layer with scale factor $r$ and then pass the concatenated feature to the next FC layer. Therefore, the attention block can perceive the scale factor and adaptively adjust the attention weights. 

In order to stabilize the training, we adopt the temperature annealing strategy introduced by \cite{chen2020dynamic}. The strategy uses a large temperature coefficient $\tau$ to make the attention weights closer to a uniform distribution, 
\begin{equation}
   \pi_k = \frac{\exp(z_k/\tau)}{\sum_{i=1}^K  \exp(z_i/\tau)},
   \label{eq:softmax}
\end{equation}
where $z_k$ is the output of the last FC layer in the attention block. With the training process, the temperature is gradually reduced to $1$, which makes (\ref{eq:softmax}) an original softmax function. 

\subsection{Scale-Aware Feature Learning (SAFL)}

In this section, we focus on formulating the SAFL part. We adopt a feedback scheme to enhance the ability of the feature learning part without increasing the number of parameters. The SAFL part can be unfolded into $T$ iterations, as shown in Fig.\,\ref{fig:flowchart}. Unlike the usual recurrent structure, each iteration accepts shallow features as part of the input and sends the output to the subsequent upsampling module.

In each iteration, we take the shallow features $F_0$ of an LR image as input to provide low-level information. First, the shallow features are refined by a hidden state $H_{t-1}$ from the previous iteration as
\begin{equation}
   L_0 = \mathopr{Conv}_{3,m}(\mathopr{Concat}(F_0, H_{t-1})),
\end{equation}
where $L_0$ is the refined feature map. 

Then, $L_0$ is sent to multiple densely connected local dense groups (LDGs). The densely connected structure enables the model to incorporate multi-level features and makes information propagation more efficient. We progressively collect the output of all LDGs through feature concatenation. To avoid the problem of high computational complexity caused by repeated concatenation, we adopt a $1\times 1$ convolution before each LDG to compress the concatenated feature and reduce the number of channels. To be precise, the output of the $d^{\mathrm{th}}$ LDG is computed as:
\begin{equation}
   L_d = \mathopr{LDG}(\mathopr{Conv}_{1, m}(\mathopr{Concat}(L_0, \cdots, L_{d-1}))),
\end{equation}
where $L_d$ denotes the output of the $d^{\mathrm{th}}$ LDG ($d>0$), and $\mathopr{LDG}$ denotes the operation of the LDG. We regard the output of the last LDG as the hidden state of the current iteration, i.e., $H_t = L_D$, where $D$ is the number of LDGs in each iteration. 

Finally, we incorporate a global skip connection in each iteration to allow the original information to bypass the sub-network. The output of the $t^{\mathrm{th}}$ iteration $F_t$ is computed as 
\begin{equation}
   F_t = H_t + \mathopr{Conv}_{5,m}(I_{\mathrm{LR}}).
\end{equation}
Collecting the output of each iteration, we obtain the input feature list $\{F_t\}_{t=1}^T$ of the following CSUM. 

\begin{figure}[t]
   \centering
   \includegraphics[width=0.85\linewidth]{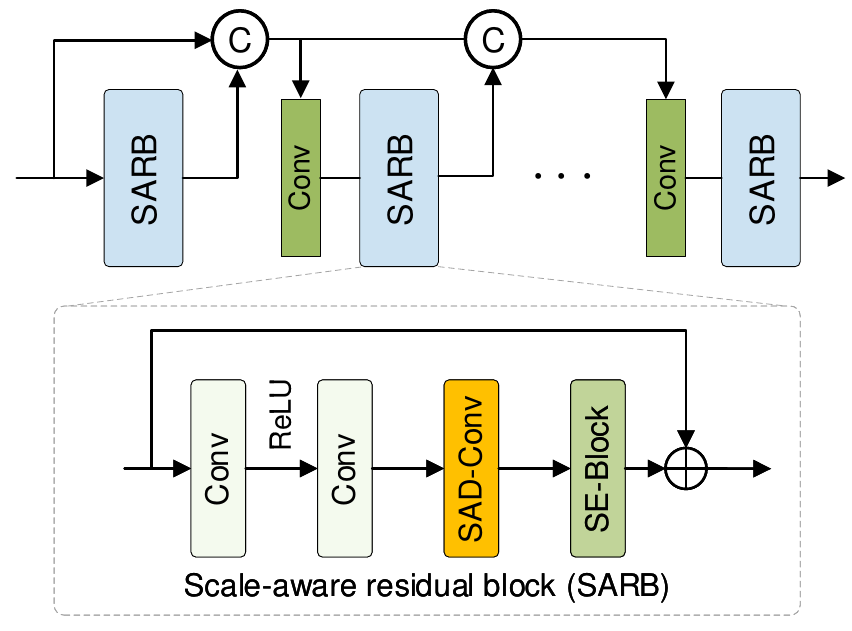}
   \caption{The architecture of the local dense group (LDG). SE-Block represents a squeeze-and-excitation block.}
   \label{fig:ldg}
   \vspace{-0.5em}
\end{figure}

\subsubsection{Local Dense Group} The architecture of the LDG is shown in Fig.\,\ref{fig:ldg}. The LDG contains $C$ scale-aware residual blocks (SARBs), which are densely connected in a similar way to LDGs. To be precise, let $R_c$ denote the output of the $c^{\mathrm{th}}$ SARB. Then,
\begin{equation}
   R_c = \mathopr{SARB}(\mathopr{Conv}_{1, m}(\mathopr{Concat}(R_0, \cdots, R_{c-1}))),
\end{equation}
where $\mathopr{SARB}$ denotes the SARB operation. Finally, the output of the LDG is the output of the last SARB, i.e., $R_C$. Next, we introduce the details of the SARB. 

\subsubsection{Scale-Aware Residual Block} First, we widen the input feature map through a $3\times 3$ convolutional layer to allow more information to pass and prevent the loss of details. The convolutional layer is followed by a ReLU activation function. Then, another $3\times 3$ convolutional layer is utilized to restore the number of channels. Next, we use an SAD-Conv to enable the residual block to perceive the scale information and dynamically adjust the parameters. Finally, to make the network adaptively emphasize informative features, we use an SE-Block \cite{hu2018squeeze} to calculate teh attention weights across channels. Let $\bm y_{\mathrm{out}}$ and $\bm y$ be the output and input feature maps of the SARB, respectively. The SARB proceeds as:
\begin{align}
   \hat{\bm y} &= \mathopr{Conv}_{3,m}(\mathopr{ReLU}(\mathopr{Conv}_{3,4m}(\bm y))),\\
   \bm y_{\mathrm{out}} &= \mathopr{SE}(\mathopr{SADConv}(\hat{\bm y})) + \bm y,
\end{align}
where $\mathopr{SE}$ and $\mathopr{SADConv}$ denote the operations of the squeeze-and-excitation block and the SAD-Conv layer, respectively. 

\subsection{Implementation Details}

Except for the last output convolutional layer, the number of features $m$ for all convolutional layers is $64$. The number of iterations $T$ in the feature learning part is set to $4$. In order to match the number of branches in the SAFL, we choose four upsampling scales in the CSUM: $r_1=1, r_2 = 2$, $r_3 = 4$, and $r_4 = 8$. In the SAFL, there are 4 LDGs in each iteration, that is, $D=4$. In each LDG, the number of SARBs is set to $4$, i.e., $C = 4$. We use weight normalization \cite{salimans2016weight} in all convolutional layers. Using the same settings as in \cite{chen2020dynamic}, we linearly reduce the temperature $\tau$ from $30$ to $1$ during the first 10 training epochs. 

We also design a lightweight version of our SADN, named SADN-L, in which $m, D, C$ are set to $32, 2, 2$, respectively. In the SADN-L, we set $T$ to $3$ and only keep three scales in the CSUM, i.e., $r_1=1, r_2 = 2, r_3 = 4$. 

\subsection{Differences from Prior Works}

\subsubsection{Comparisons with the LIIF} Although the LIIF \cite{chen2021learning} incorporates the local implicit function for SR, there are still distinct differences between our proposed MBLIF and LIIF. First, the LIIF only uses a single-scale feature map to construct the implicit function while we make full use of multi-scale features and obtain better performance. Second, the LIIF only takes a nearest neighbor latent code of the query coordinate as input while we use a set of latent codes near the query coordinate to increase the receptive field. Third, the LIIF needs a mandatory local ensemble strategy to obtain a continuous image representation, which eliminates artifacts by averaging multiple predictions. Our MBLIF can directly obtain the continuous representation due to the bilinearity of the local implicit feature function.

\subsubsection{Comparisons with the MetaSR} First, the MetaSR \cite{Hu2019} dynamically predicts the weights of filters for upsampling, but the weights in the feature extraction part are constant for all scale factors. Conversely, the feature extractor in our SADN can adaptively adjust parameters to conduct SR tasks with various scale factors. Second, the weight prediction in MetaSR is performed through an additional network while the weight prediction in our SADN is implemented by an attention block, which is more efficient. 

\subsubsection{Comparisons with the RDN} First, the RDN \cite{Zhang2018d} is a very large model, and can only process a single integer scale factor. Second, the RDN introduces residual dense connections only locally while we further adopt dense connections globally, which make the hierarchical information propagate more efficiently in the network. Third, the RDN concatenates all the features of skip connections and directly inputs the concatenated feature into the next convolutional layer, resulting in high computational complexity. We propose using a $1\times 1$ convolution to compress the concatenated feature, which can greatly reduce the number of parameters and improve computational efficiency.

\section{Experiments}

In this section, we first introduce the datasets, metrics, and training details. Then, we compare our model with the SOTA continuous scale and fixed scale SR methods. Finally, we investigate the effectiveness of each component through ablation experiments.

\begin{table*}[htbp]
    \centering
    \renewcommand\arraystretch{0.8}
    \caption{PSNR Results for Continuous-Scale SR on the B100 Dataset. The Best Performance is Shown in \textbf{Bold} and the Second Best \underline{Underlined}. * Indicates Our Proposals.}
    \setlength{\tabcolsep}{1.9mm}{
     \begin{tabular}{|l|ccccccccccc|cccc|}
      \hline
      \multicolumn{1}{|c|}{\multirow{2}[4]{*}{Methods}} & \multicolumn{11}{c|}{In-distribution}     & \multicolumn{4}{c|}{Out-of-distribution} \bigstrut\\
 \cline{2-16}       & $\times$1.1 & $\times$1.4 & $\times$1.6 & $\times$1.8 & $\times$2 & $\times$2.3 & $\times$2.6 & $\times$3 & $\times$3.4 & $\times$3.9 & $\times$4 & $\times$6 & $\times$8 & $\times$10 & $\times$20 \bigstrut\\
      \hline
      \hline
      Bicubic & 36.59  & 32.96  & 31.50  & 30.42  & 29.60  & 28.64  & 27.92  & 27.20  & 26.61  & 26.05  & 25.96  & 24.53  & 23.68  & 23.03  & 21.12 \bigstrut[t]\\
      OSM-EDSR-B \cite{behjati2021overnet} & 41.60  & 36.67  & 34.64  & 33.22  & 32.12  & 30.90  & 29.98  & 29.05  & 28.35  & 27.65  & 27.54  & 25.79  & 24.75  & 24.02  & 21.94  \\
      Meta-EDSR-B \cite{Hu2019} & 42.43  & 36.75  & 34.72  & 33.27  & 32.15  & 30.92  & 30.01  & 29.09  & 28.37  & 27.66  & 27.54  & 25.75  & 24.70  & 23.97  & 21.87  \\
      LIIF-EDSR-B \cite{chen2021learning} & 42.53  & 36.77  & 34.73  & 33.28  & 32.17  & 30.94  & 30.03  & 29.10  & 28.40  & 27.71  & 27.60  & 25.85  & 24.79  & 24.05  & 21.93  \\
      CSUM-EDSR-B* & 42.78  & 36.80  &34.74  & 33.30  & 32.18  & 30.97  & 30.06  & 29.12  & 28.42  & 27.73  & 27.61  & 25.86  & 24.81  & 24.07  & 21.96 \bigstrut[b]\\
      \hline
      Bi-RDN \cite{Zhang2018d} & 39.85  & 36.39  & 34.61  & 33.20  & 32.13  & 30.96  & 30.04  & 29.15  & 28.43  & 27.75  & 27.63  & 25.92  & 24.83  & 24.10  & 22.00  \bigstrut[t]\\
      OSM-RDN \cite{behjati2021overnet} & 42.63  & 36.78  & 34.79  & 33.37  & 32.27  & 31.04  & 30.12  & 29.19  & 28.50  & 27.81  & 27.70  & -- & -- & -- & -- \\
      Meta-RDN \cite{Hu2019} & 42.74  & 36.88  & 34.87  & 33.43  & 32.31  & 31.08  & 30.19  & 29.26  & 28.54  & 27.84  & 27.71  & 25.90  & 24.83  & 24.06  & 21.94  \\
      LIIF-RDN \cite{chen2021learning} & 42.83  & 36.92  & 34.89  & 33.43  & 32.31  & 31.09  & 30.19  & 29.27  & 28.56  & 27.86  & \underline{27.75} & 25.98  & 24.92  & 24.15  & 22.02  \\
      CSUM-RDN* & \underline{43.03}  & \underline{36.95}  & \underline{34.90}  & \underline{33.45}  & \underline{32.33}  & \underline{31.13}  & \underline{30.20}  & \underline{29.29}  & \underline{28.57}  & \underline{27.87}  & \underline{27.75}  & \underline{25.99}  & \underline{24.94}  & \underline{24.17}  & \textbf{22.04} \bigstrut[b]\\
      \hline
      SADN-L* & 42.86  & 36.85  & 34.78  & 33.33  & 32.20  & 31.00  & 30.08  & 29.15  & 28.45  & 27.75  & 27.63  & 25.90  & 24.85  & 24.11  & 21.99  \bigstrut[t]\\
      SADN* & \textbf{43.10} & \textbf{36.98} & \textbf{34.93} & \textbf{33.49} & \textbf{32.37} & \textbf{31.17} & \textbf{30.26} & \textbf{29.32} & \textbf{28.62} & \textbf{27.92} & \textbf{27.78} & \textbf{26.03} & \textbf{24.98} & \textbf{24.19} & \underline{22.03}  \bigstrut[b]\\
      \hline
      \end{tabular}%
 
    }
    \label{tab:scale-arbitrary-results}
\end{table*}

\subsection{Datasets and Metrics}
We use DIV2K\cite{agustsson2017ntire} containing 800 high-resolution images as our training dataset. For testing, we evaluate SR models on five standard benchmark datasets: Set5\cite{bevilacqua2012low}, Set14\cite{zeyde2010single}, B100\cite{martin2001database}, Manga109\cite{matsui2017sketch}, and Urban100\cite{huang2015single}. Following the settings of previous works, we only consider the PSNR and SSIM\cite{wang2004image} of the Y channel of the transformed YCbCr space. Following previous works, we use bicubic downsampling as the standard degradation model for generating LR images from their HR counterparts.

\begin{table*}[ht]
    \centering
    \caption{Average Results of PSNR({\rm dB}) and SSIM for Integer Scale Factors $\times 2$, $\times 3$ and $\times 4$. The Best Performance is Shown in \textbf{Bold} and the Second Best \underline{Underlined}. * Indicates Our Proposals.}
    \renewcommand\arraystretch{0.8}
    \setlength{\tabcolsep}{1.1mm}{
     \begin{tabular}{|lc|ccc|ccc|ccc|ccc|ccc|}
      \hline
      \multirow{2}[4]{*}{Methods} & \multirow{2}[4]{*}{Metric} & \multicolumn{3}{c|}{Set5} & \multicolumn{3}{c|}{Set14} & \multicolumn{3}{c|}{Urban100} & \multicolumn{3}{c|}{BSD100} & \multicolumn{3}{c|}{Manga109} \bigstrut\\
 \cline{3-17}       &   & $\times$2 & $\times$3 & $\times$4 & $\times$2 & $\times$3 & $\times$4 & $\times$2 & $\times$3 & $\times$4 & $\times$2 & $\times$3 & $\times$4 & $\times$2 & $\times$3 & $\times$4 \bigstrut\\
      \hline
      \hline
      \multirow{2}[0]{*}{Bicubic} & PSNR & 33.66  & 30.39  & 28.42  & 30.24  & 27.55  & 26.00  & 26.66  & 24.47  & 23.15  & 29.56  & 27.21  & 25.96  & 30.80  & 26.95  & 24.89  \bigstrut[t]\\
        & SSIM & 0.9299  & 0.8682  & 0.8104  & 0.8688  & 0.7742  & 0.7227  & 0.8410  & 0.7370  & 0.6585  & 0.8431  & 0.7385  & 0.6675  & 0.9339  & 0.8556  & 0.7866  \bigstrut[b]\\
      \hline
      \multirow{2}[0]{*}{CFSRCNN\cite{tian2020coarse}} & PSNR & 37.79  & 34.24  & 32.06  & 33.51  & 30.27  & 28.57  & 32.07  & 28.04  & 26.03  & 32.11  & 29.03  & 27.53  & -- & -- & -- \bigstrut[t]\\
        & SSIM & 0.9591  & 0.9256  & 0.8920  & 0.9165  & 0.8410  & 0.7800  & 0.9273  & 0.8496  & 0.7824  & 0.8999  & 0.8035  & 0.7333  & -- & -- & -- \bigstrut[b]\\
      \hline
      \multirow{2}[0]{*}{SRFBN\cite{Li2019c}} & PSNR & 38.11  & 34.70  & 32.47  & 33.82  & 30.51  & 28.81  & 32.62  & 28.73  & 26.60  & 32.29  & 29.24  & 27.72  & 39.08  & 34.18  & 31.15  \bigstrut[t]\\
        & SSIM & 0.9609  & 0.9292  & 0.8983  & 0.9196  & 0.8461  & 0.7868  & 0.9328  & 0.8641  & 0.8015  & 0.9010  & 0.8084  & 0.7409  & 0.9779  & 0.9481  & 0.9160  \bigstrut[b]\\
      \hline
      \multirow{2}[0]{*}{ISNR\cite{liu2021iterative}} & PSNR & 38.20  & 34.68  & 32.55  & 33.84  & 30.60  & 28.79  & 32.96  & 28.83  & 26.64  & 32.35  & 29.25  & 27.74  & 39.20  & 34.19  & 31.16  \bigstrut[t]\\
        & SSIM & 0.9613  & 0.9294  & 0.8992  & 0.9199  & 0.8475  & 0.7872  & 0.9357  & 0.8666  & 0.8033  & 0.9019  & 0.8096  & 0.7422  & 0.9781  & 0.9487  & 0.9166  \bigstrut[b]\\
      \hline
      \multirow{2}[0]{*}{RDN\cite{Zhang2018d}} & PSNR & 38.24  & 34.71  & 32.47  & 34.01  & 30.57  & 28.81  & 32.96  & 28.80  & 26.61  & 32.34  & 29.26  & 27.72  & 39.18  & 34.13  & 31.00  \bigstrut[t]\\
        & SSIM & 0.9614  & 0.9296  & 0.8990  & 0.9212  & 0.8468  & 0.7871  & 0.9360  & 0.8650  & 0.8030  & 0.9017  & 0.8093  & 0.7419  & 0.9780  & 0.9484  & 0.9151  \bigstrut[b]\\
      \hline
      \multirow{2}[0]{*}{SAN\cite{dai2019second}} & PSNR & \textbf{38.31} & 34.75  & 32.64  & 34.07  & 30.59  & \underline{28.92}  & 33.10  & 28.93  & 26.79  & \textbf{32.42} & \underline{29.33}  & \underline{27.78}  & 39.32  & 34.30  & 31.18  \bigstrut[t]\\
        & SSIM & \textbf{0.9620} & \underline{0.9300}  & 0.9003  & 0.9213  & 0.8476  & 0.7888  & \underline{0.9370}  & 0.8671  & 0.8068  & \textbf{0.9028} & \underline{0.8112}  & 0.7436  & \textbf{0.9792} & 0.9494  & 0.9169  \bigstrut[b]\\
      \hline
      \hline
      \multirow{2}[0]{*}{OverNet\cite{behjati2021overnet}} & PSNR & 38.00  & 34.30  & 32.20  & 33.61  & 30.35  & 28.65  & 31.97  & 28.22  & 26.08  & 32.13  & 29.08  & 27.57  & 38.81  & 33.80  & 30.64  \bigstrut[t]\\
        & SSIM & 0.9602  & 0.9262  & 0.8951  & 0.9180  & 0.8424  & 0.7829  & 0.9268  & 0.8529  & 0.7853  & 0.8988  & 0.8050  & 0.7363  & 0.9772  & 0.9454  & 0.9089  \bigstrut[b]\\
      \hline
      \multirow{2}[0]{*}{LIIF-RDN\cite{chen2021learning}} & PSNR & 38.16  & 34.67  & 32.49  & 33.96  & 30.53  & 28.80  & 32.87  & 28.82  & 26.68  & 32.31  & 29.27  & 27.75  & 39.24  & 34.20  & 31.21  \bigstrut[t]\\
        & SSIM & 0.9610  & 0.9291  & 0.8988  & 0.9209  & 0.8470  & 0.7874  & 0.9352  & 0.8663  & 0.8040  & 0.9010  & 0.8100  & 0.7421  & 0.9781  & 0.9488  & 0.9170  \bigstrut[b]\\
      \hline
      \multirow{2}[0]{*}{Meta-RDN\cite{Hu2019}} & PSNR & 38.18  & 34.67  & 32.42  & 33.95  & 30.55  & 28.77  & 32.86  & 28.80  & 26.53  & 32.31  & 29.26  & 27.71  & 39.24  & 34.33  & 31.17  \bigstrut[t]\\
      & SSIM & 0.9612  & 0.9291  & 0.8979  & 0.9208  & 0.8468  & 0.7867  & 0.9350  & 0.8655  & 0.8007  & 0.9015  & 0.8097  & 0.7412  & 0.9781  & 0.9488  & 0.9159  \bigstrut[b]\\
      \hline
      \multirow{2}[0]{*}{SADN-L*} & PSNR & 38.05  & 34.54  & 32.32  & 33.76  & 30.42  & 28.72  & 32.35  & 28.37  & 26.30  & 32.20  & 29.15  & 27.63  & 38.90  & 33.76  & 30.65  \bigstrut[t]\\
       & SSIM & 0.9607  & 0.9281  & 0.8968  & 0.9190  & 0.8435  & 0.7843  & 0.9298  & 0.8562  & 0.7924  & 0.8999  & 0.8063  & 0.7382  & 0.9776  & 0.9461  & 0.9115  \bigstrut[b]\\
      \hline
      \multirow{2}[0]{*}{SADN*} & PSNR & 38.24  & \underline{34.78}  & \underline{32.69}  & \underline{34.10}  & \underline{30.65}  & 28.91  & \underline{33.16}  & \underline{29.07}  & \underline{26.90}  & 32.37  & \underline{29.33}  & \underline{27.78}  & \underline{39.42}  & \underline{34.47}  & \underline{31.36}  \bigstrut[t]\\
      & SSIM & 0.9614  & \underline{0.9300}  & \underline{0.9005}  & \underline{0.9216}  & \underline{0.8478}  & \underline{0.7896}  & 0.9369  & \underline{0.8696}  & \underline{0.8094}  & 0.9017  & 0.8108  & \underline{0.7441}  & 0.9786  & \underline{0.9502}  & \underline{0.9196}  \bigstrut[b]\\
      \hline
      \multirow{2}[0]{*}{SADN+*} & PSNR & \underline{38.30}  & \textbf{34.88} & \textbf{32.77} & \textbf{34.16} & \textbf{30.71} & \textbf{28.99} & \textbf{33.32} & \textbf{29.23} & \textbf{27.05} & \underline{32.40}  & \textbf{29.36} & \textbf{27.82} & \textbf{39.54} & \textbf{34.66} & \textbf{31.58 } \bigstrut[t]\\
        & SSIM & \underline{0.9616}  & \textbf{0.9306} & \textbf{0.9016} & \textbf{0.9218} & \textbf{0.8486} & \textbf{0.7909} & \textbf{0.9379} & \textbf{0.8714} & \textbf{0.8121} & \underline{0.9021}  & \textbf{0.8115} & \textbf{0.7450} & \underline{0.9788}  & \textbf{0.9511} & \textbf{0.9212 } \bigstrut[b]\\
      \hline
      \end{tabular}%
    }
    \label{tab:scale-fixed-results}
\end{table*}

\subsection{Training Details}
In the training phase, we randomly crop LR patches with a size of $48\times 48$ as input. We randomly flip images vertically or horizontally and randomly rote them $90^{\circ}$ for data augmentation. We train our model for 1000 epochs with a mini-batch size of 16, and each epoch contains 1000 iterations. The initial learning is $1\times 10^{-4}$ and decreases by half every 200 epochs. The scale factors of each batch of images are the same, and the scale factors in the training phase are uniformly distributed from 1 to 4. Our model is implemented using the PyTorch framework and trained on an Nvidia GeForce RTX 3090 GPU.

\subsection{Comparison with Continuous-Scale SR Methods}
\begin{figure*}[htbp]
  \centering
  \includegraphics[width=0.95\textwidth]{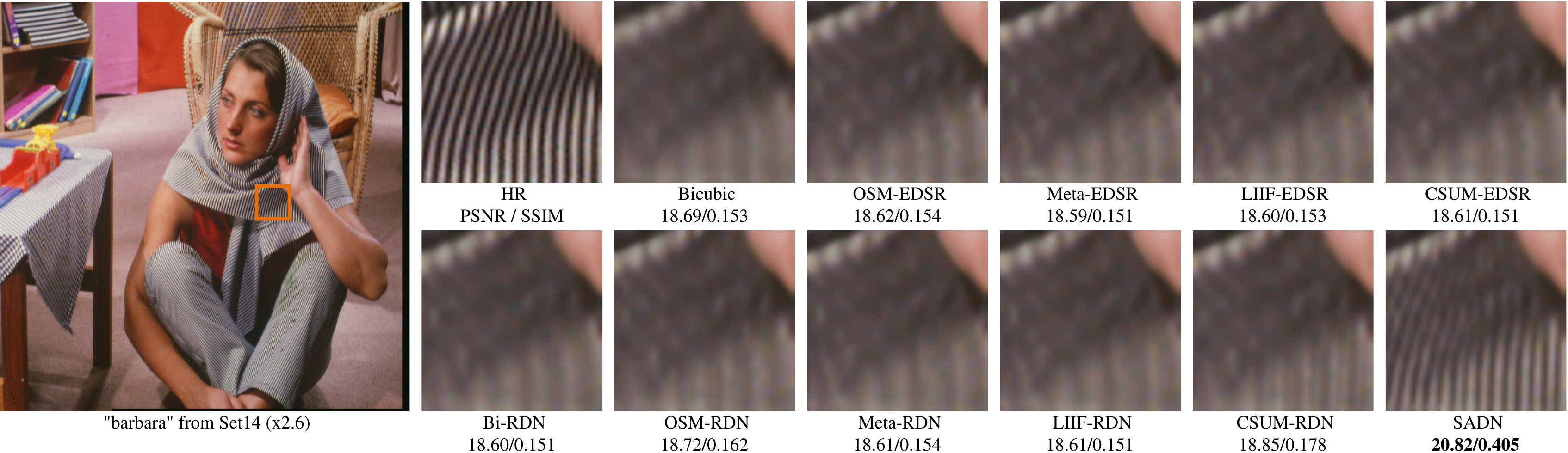}
  \vspace{5pt}\\
  \includegraphics[width=0.95\textwidth]{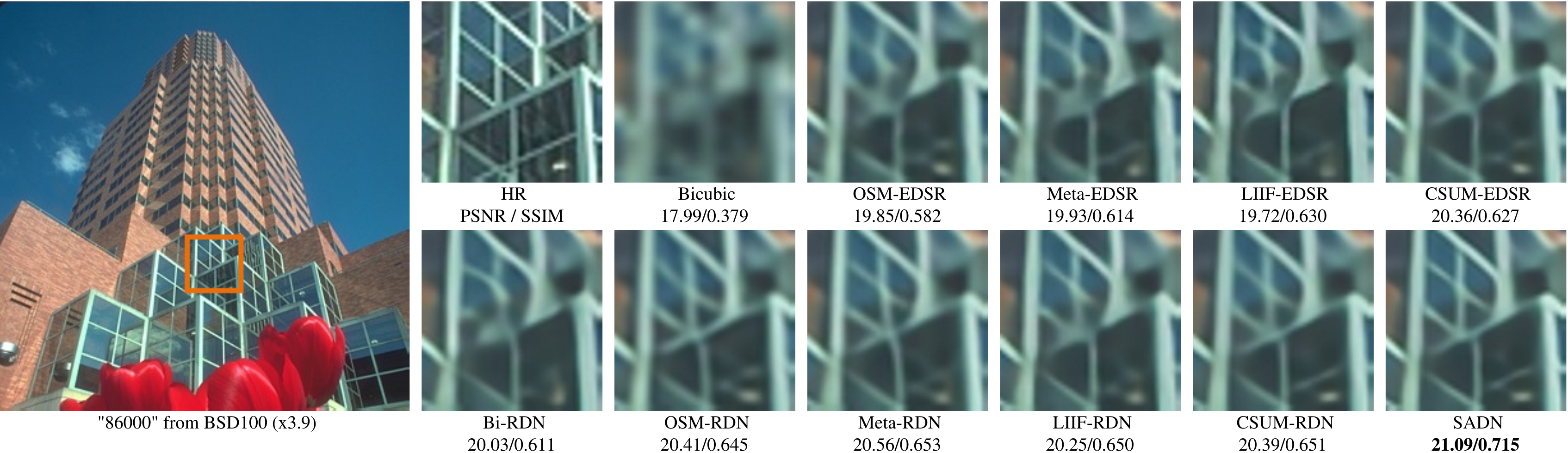}
  \caption{Visual comparisons between continuous-scale SR methods. The top two rows are results with a scale factor of $\times 2.6$ and the bottom two rows are results with a scale factor of $\times 3.9$.}
  \label{fig:mainrslt_Set14_x26}
\end{figure*}

\begin{figure*}[tbp]
  \centering
  \includegraphics[width=0.95\linewidth]{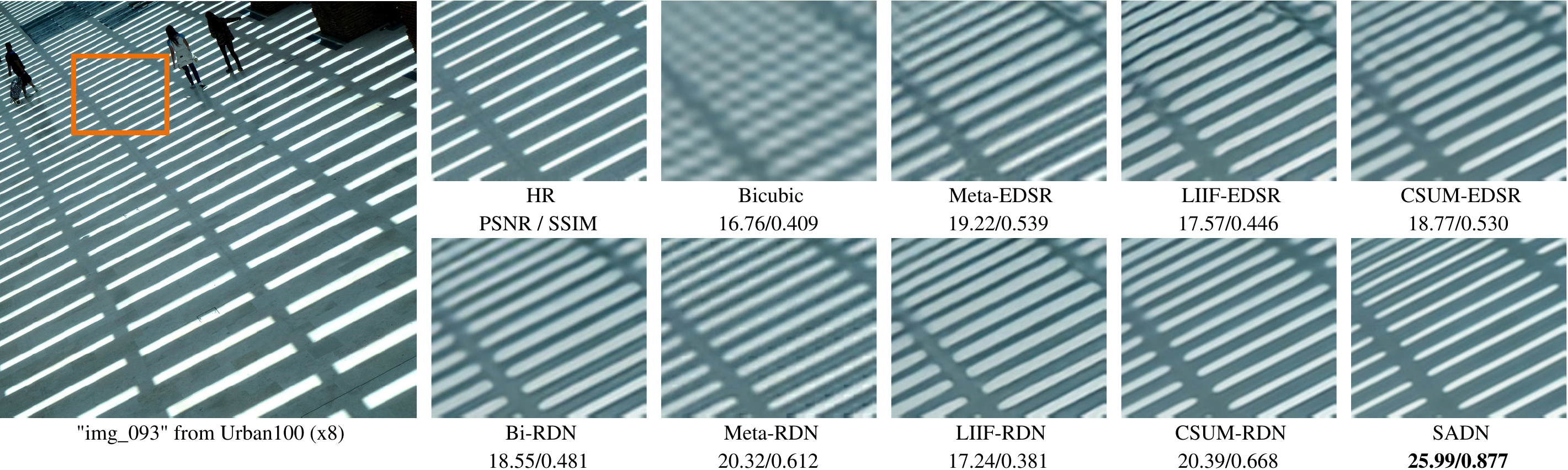}
  \vspace{5pt}\\
  \includegraphics[width=0.95\linewidth]{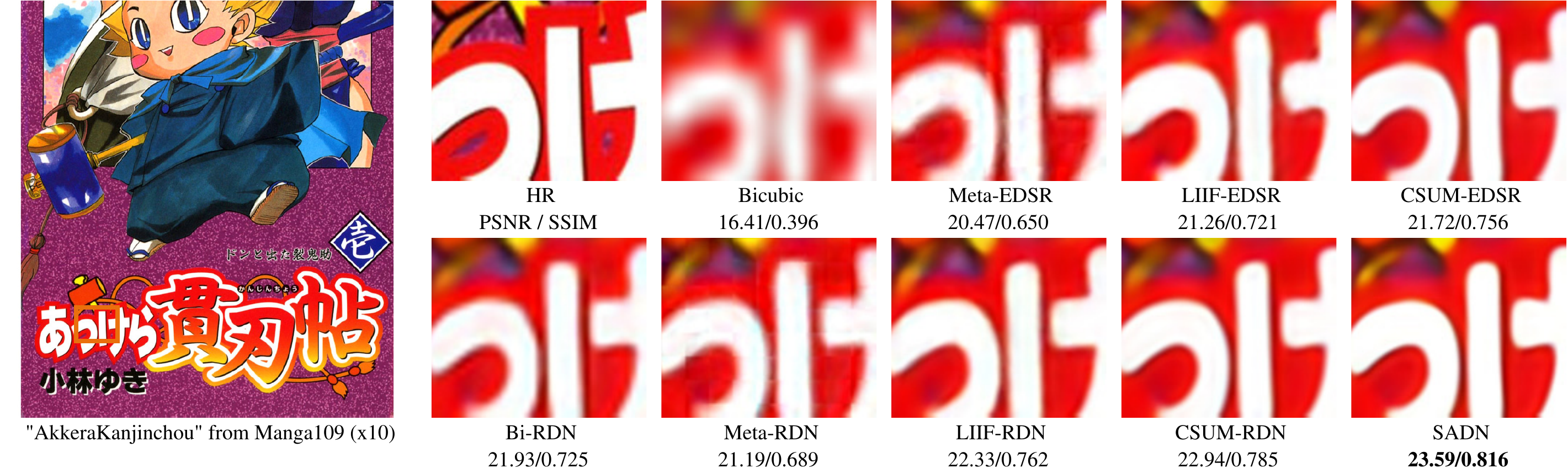}
  \caption{Visual comparisons with out-of-distribution integer scale factors. The top two rows are results with a scale factor of $\times 8$ and the bottom two rows are results with a scale factor of $\times 10$.}
  \label{fig:mainrslt_large_scale}
\end{figure*}

\begin{figure*}[tbp]
  \centering
  \includegraphics[width=0.95\linewidth]{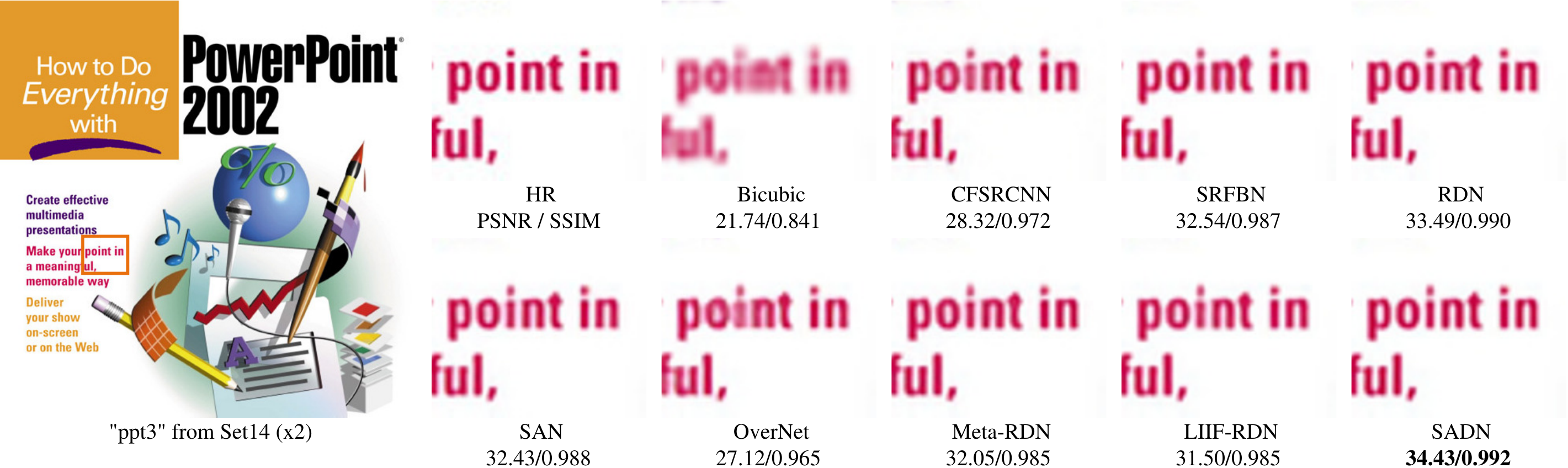}
  \vspace{5pt}\\
  \includegraphics[width=0.95\linewidth]{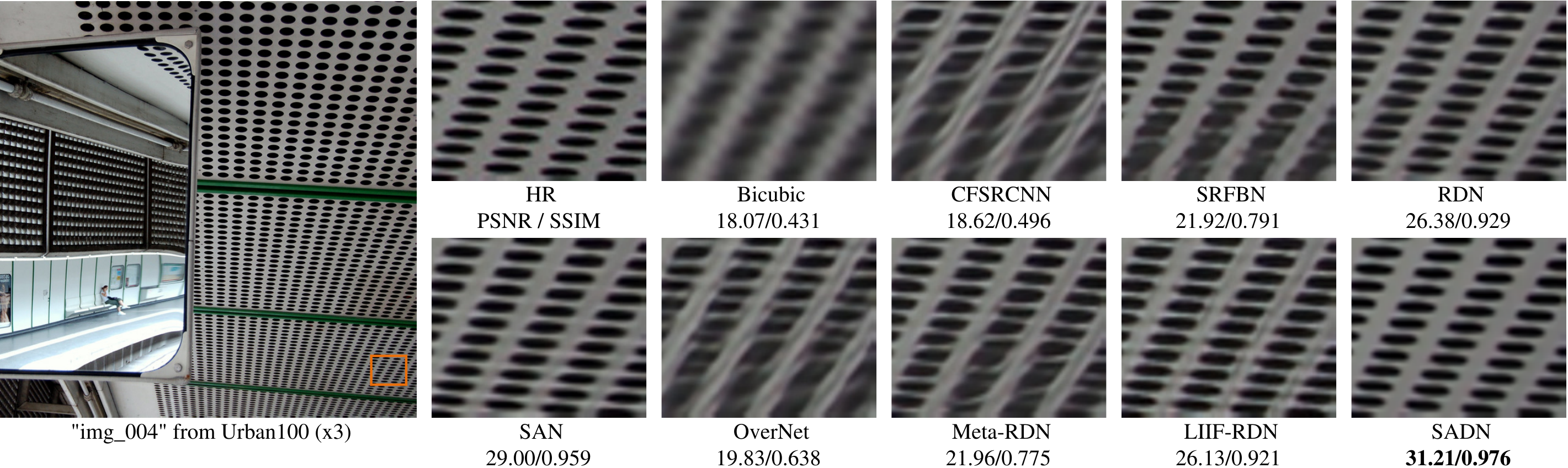}
  \vspace{5pt}\\
  \includegraphics[width=0.95\linewidth]{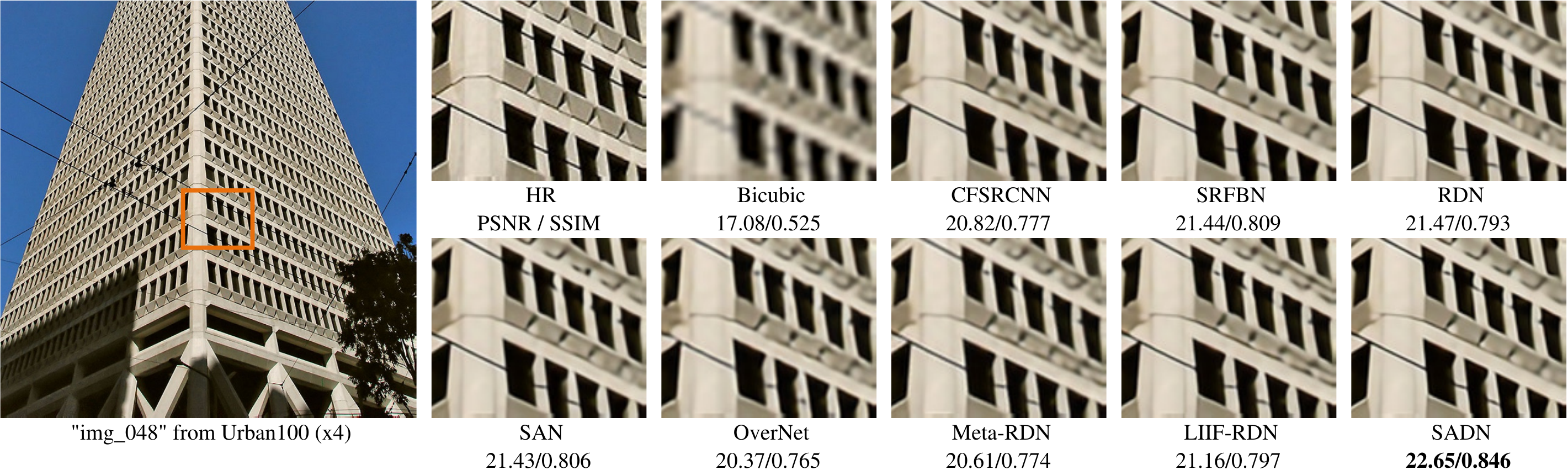}
  \caption{Visual comparisons between SOTA fixed-scale and continuous-scale SR methods. The top two rows are results with a scale factor of $\times$2, the middle two rows are results with a scale factor of $\times$3 and the bottom two rows are results with a scale factor of $\times$4.}
  \label{fig:mainrslt_integer_scales}
\end{figure*}

We take bicubic interpolation as the baseline and compare our proposed CSUM with three continuous-scale upsampling modules. The meta-upsampling module proposed in MetaSR \cite{Hu2019} is a pioneering method for arbitrary-scale upsampling. The LIIF \cite{chen2021learning} is an SOTA continuous-scale SR method based on implicit functions. The OSM \cite{behjati2021overnet}, based on overscaling and interpolation, can achieve any scale SR within a certain range. For a fair comparison, we replace the original upsampling layer of EDSR-baseline (EDSR-B) \cite{Lim2017} and the RDN \cite{Zhang2018d} with the OSM, the meta-upsampling module, the LIIF, and our proposed CSUM. Therefore, we obtain six different model combinations, namely, OSM-EDSR-B, Meta-EDSR-B, LIIF-EDSR-B, CSUM-EDSR-B, OSM-RDN, Meta-RDN, LIIF-RDN, and CSUM-RDN. Besides, we define a new baseline by resampling the output of a fixed-scale SR model to the target size. Specifically, we adopt the RDN model with a fixed scaling factor of $\times$4. For scaling factor $r$, the input LR image is first upscaled by the RDN model and then resampled by bicubic interpolation with a scale factor of $r/4$. This baseline is called the Bi-RDN. We also add our proposed SADN and the lightweight version SADN-L for comparison.

The results of the OSM and MetaSR are from our implementation, and the results of the LIIF are obtained by using the official code and pre-trained models. The CSUM-EDSR and CSUM-RDN are retrained from scratch. In addition to the scale factors in the training distribution ($\times 1$ -- $\times$4), we also evaluate models on large-scale factors out of the training distribution to test the generalization performance. The quantitative results are shown in Table\,\ref{tab:scale-arbitrary-results}. The best results of all comparisons are bolded, and the second best results are underlined.

As the table shows, our proposed CSUM-EDSR-B obtains the best results among all EDSR-B-based methods, and CSUM-RDN obtains the best results among all RDN-based methods. In particular, the proposed CSUM has made significant progress on small scale factors, such as an increase of $0.25$ dB with a scale factor $\times 1.1$. The performance of CSUM-EDSR-B and CSUM-RDN demonstrates that the proposed CSUM can replace the upsampling layer of previous SR networks and obtain satisfactory results. Moreover, benefiting from the scale-aware feature extractor, our proposed SADN obtains the best object metrics on almost all scale factors.

We compare the visual results of the continuous-scale SR methods in Fig.\,\ref{fig:mainrslt_Set14_x26}. Furthermore, Fig.\,\ref{fig:mainrslt_large_scale} shows the visual comparisons of large-scale factors out of the training distribution. Our proposed SADN performs better at restoring clear details and sharp edges. Besides, CSUM-EDSR-B and CSUM-RDN obtain more convincing results among all EDSR-B-based and RDN-based methods, respectively. In summary, our proposed upsampling module combined with the scale-aware feature extractor can achieve better visual performance.

\subsection{Comparison with the SOTA Methods}

We compare our proposed SADN with SOTA fixed-scale SR methods, including CFSRCNN \cite{tian2020coarse}, SRFBN \cite{Li2019c}, RDN \cite{Zhang2018d}, ISNR \cite{liu2021iterative}, and SAN \cite{dai2019second}; and continuous-scale SR methods, including OverNet \cite{behjati2021overnet}, Meta-RDN \cite{Hu2019} and LIIF-RDN \cite{chen2021learning}, on five benchmark datasets. The results of RDN, SRFBN, and SAN are obtained by retraining on each scale factor. The results of the continuous-scale comparative methods are obtained using a single model without fine-tuning. We also use the self-ensemble \cite{timofte2016seven} strategy to further improve the performance of the SADN, and the method is named SADN+.

Table\,\ref{tab:scale-fixed-results} shows the quantitative results with scale factors of $\times$2, $\times$3, and $\times$4. Our proposed SADN achieves the best performance on most datasets and scales. The results demonstrate that our proposed SADN has made great progress compared to the existing continuous-scale SR methods and obtains comparable and even better results with the SOTA fixed-scale methods, such as the SAN. Compared with the lightweight networks CFSRCNN and OverNet, our SADN-L also performs better on objective metrics. Thanks to the CSUM, our proposed SADN and SADN-L only need to be trained once and save a single model while the fixed-scale methods need to be trained several times and save a model for each scale factor. Therefore, the SADN not only greatly saves training time and model storage space compared to fixed-scale SR models but also bridges the performance gap between the continuous-scale SR models and the fixed-scale SR models.

Fig.\,\ref{fig:mainrslt_integer_scales} shows the visual results with scale factors of $\times$2, $\times$3, and $\times$4. In general, the SADN can obtain more realistic SR images. For the SR results of ``ppt3'' from Set14, our SADN can clearly recover all letters, especially the letter ``i'' and the letter ``t''. For ``img\_004'' from Urban100, the proposed SADN produces circles that are very close to the ground truth. The good visual results are attributable to the fact that our proposed SADN makes full use of multi-level features and a larger range of latent codes to achieve reconstruction, so more realistic SR images can be obtained at different scales. In addition, scale-aware feature extraction also ensures that good results can be obtained at multiple scale factors.

\subsection{Network Parameters}
\begin{figure}[t]
  \centering
  \includegraphics[width=0.9\linewidth]{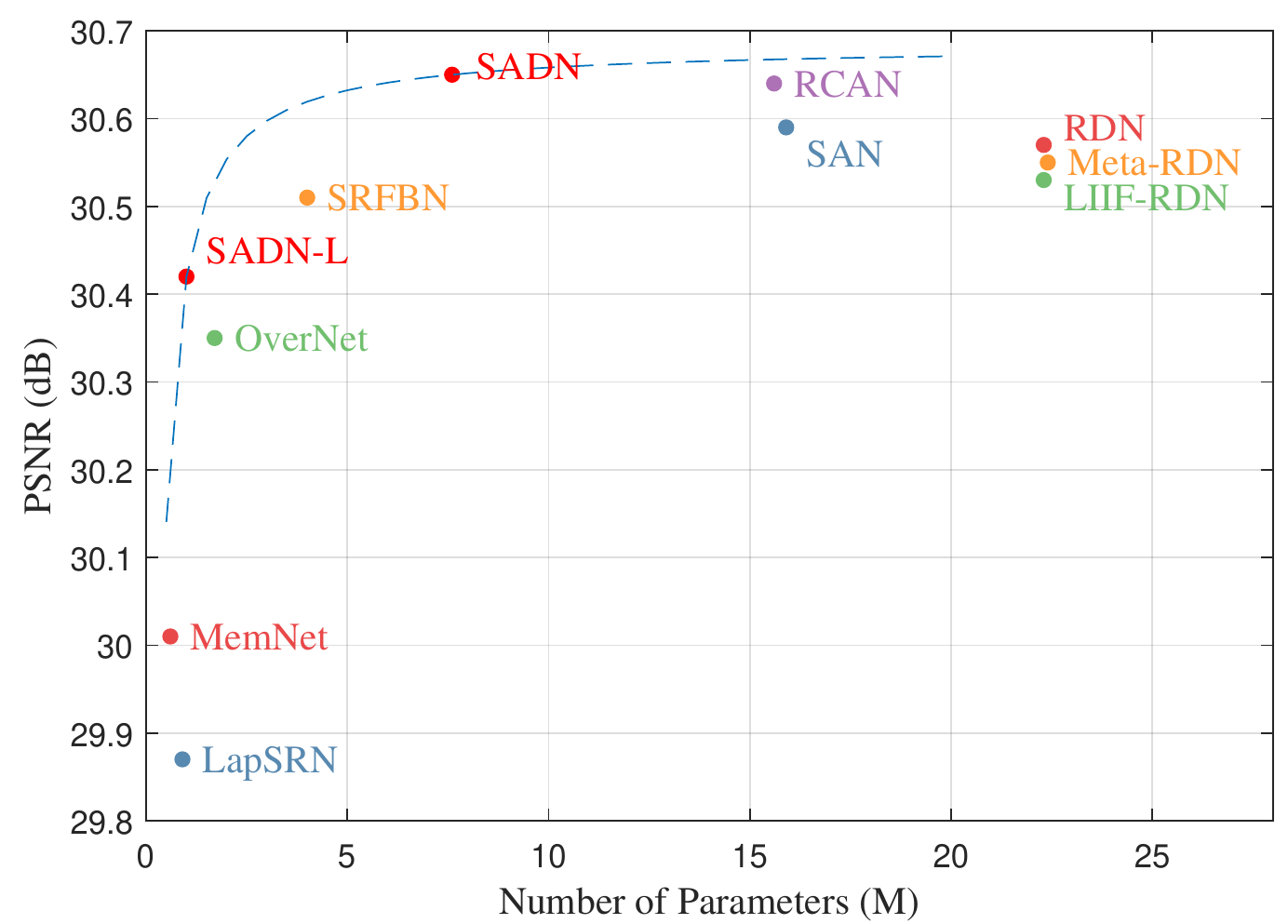}
  \caption{Performance vs. number of parameters. The results are evaluated on the Set14 dataset with a scale factor of $\times$3. The red points represent our proposed networks.}
  \label{fig:parameter}
  \vspace{-0.5em}
\end{figure}

We compare the network parameters and model performance of our proposed SADN with nine comparative methods, including LapSRN\cite{Lai2017deep}, MemNet\cite{tai2017memnet}, OverNet\cite{behjati2021overnet}, SRFBN\cite{Li2019c}, RCAN\cite{zhang2018image}, SAN\cite{dai2019second}, RDN\cite{Zhang2018d}, LIIF-RDN \cite{chen2021learning}, and Meta-RDN \cite{Hu2019}. The comparison results are shown in Fig.\,\ref{fig:parameter}.

The proposed SADN achieves the best performance and uses only 7.6\,M parameters, which is 1/3 of the number of parameters of MetaSR and LIIF. Furthermore, our proposed lightweight model SADN-L has fewer than 1\,M parameters, and it outperforms OverNet with 1.7\,M parameters. Thanks to the feedback structure, our method can balance the number of network parameters and model performance well.

\begin{table}[t]
  \caption{Average Running Time Comparison on Set14 Dataset with a Scale Factor of $\times$4 on an NVIDIA 3090 GPU.  * Indicates Our Proposals.}
  \renewcommand\arraystretch{0.9}
  \centering
  \begin{tabular}{|l|ccc|}
    \hline
    Method                            & \multicolumn{1}{l}{Parameters (M)} & \multicolumn{1}{l}{Running Time (s)} & \multicolumn{1}{l|}{PSNR (dB)} \bigstrut \\
    \hline\hline
    OverNet \cite{behjati2021overnet} & 1.7                                & 0.0173                               & 28.65 \bigstrut[t]                       \\
    SADN-L*                           & 1.0                                & 0.0232                               & 28.72                                    \\
    RDN \cite{Zhang2018d}             & 22.3                               & 0.0426                               & 28.81                                    \\
    CSUM-RDN*                         & 22.7                               & 0.0628                               & 28.82                                    \\
    Meta-RDN \cite{Hu2019}            & 22.4                               & 0.0666                               & 28.77                                    \\
    SADN*                             & 7.6                                & 0.0881                               & 28.91                                    \\
    SAN \cite{dai2019second}          & 15.9                               & 0.1118                               & 28.92                                    \\
    LIIF-RDN \cite{chen2021learning}  & 22.3                               & 0.1508                               & 28.80                                    \\
    SRFBN \cite{Li2019c}              & 4.0                                & 0.3324                               & 28.81  \bigstrut[b]                      \\
    \hline
  \end{tabular}%
  \label{tab:runtime}
\end{table}

\subsection{Running Time}
We compare the running time of our proposed methods with six methods, OverNet \cite{behjati2021overnet}, RDN \cite{Zhang2018d}, Meta-RDN \cite{Hu2019}, SAN \cite{dai2019second}, LIIF-RDN \cite{chen2021learning}, and SRFBN \cite{Li2019c}, on the Set14 dataset with a scale factor of $\times$4. The running time of all comparative methods is evaluated on an NVIDIA 3090 GPU. The comparison results are shown in Table\,\ref{tab:runtime}.

First, comparing the CSUM-RDN, Meta-RDN, and LIIF-RDN, we find that the CSUM has a running time that is only $0.02\,\mathrm{s}$ longer than that of the baseline RDN method while the LIIF takes $0.11\,\mathrm{s}$ longer. The results demonstrate that our proposed CSUM is a highly efficient upsampling module that runs 5.5$\times$ faster than the LIIF on the Set14 dataset. Moreover, compared to the SOTA SAN, our proposed SADN runs 1.3$\times$ faster and achieves comparable performance. The results show that our proposed SAFL part can extract powerful high-level features with low computational costs.

\section{Discussion}
In this section, we first investigate the random scale training strategy. Then, we present ablation studies of the feature extractor and the CSUM. Finally, we investigate the parameter settings.

\subsection{Training with a Single Scale} To investigate the differences and relationships between the SR tasks of various scale factors, below, we test the generalization performance of a model trained on a single scale factor. We train the proposed SADN with three single scales ($\times $2, $\times $3, and $\times $4) and obtain three models, named SADN-x2-only, SADN-x3-only, and SADN-x4-only, respectively. Then, the three models are tested across scales. We compare the three models with the SADN, which is trained with randomly sampled scale factors.

\begin{figure}[tbp]
  \centering
  \includegraphics[width=0.8\linewidth]{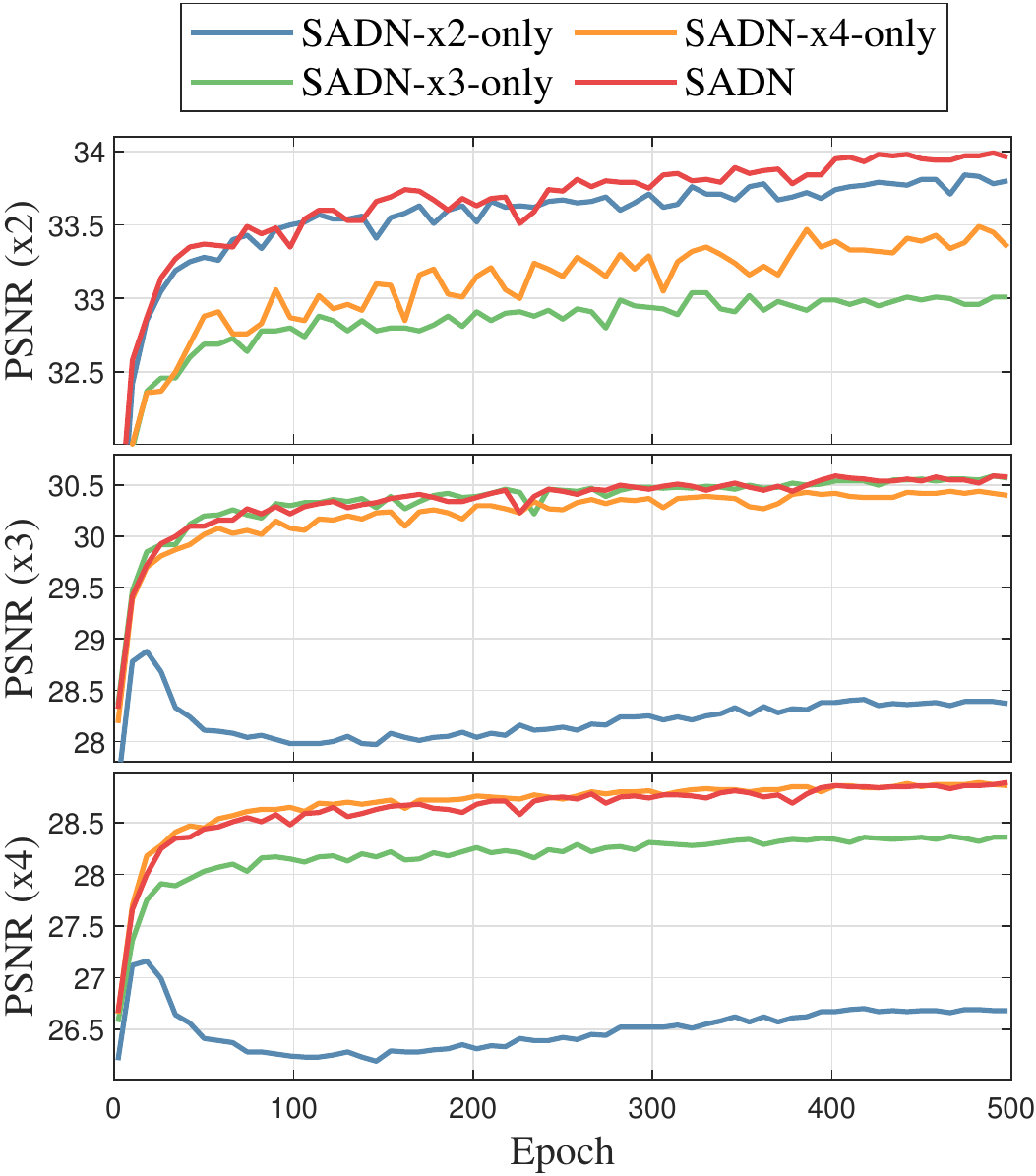}
  \caption{Investigation of the generalization performance across scales.}
  \label{fig:cross_scale}
  \vspace{-1em}
\end{figure}

Fig.\,\ref{fig:cross_scale} shows the comparison results. We have two observations. First, the model trained with a single scale factor can only obtain satisfactory results on the trained scale factor while the SADN obtains the best performance on all scale factors. This observation indicates that there are essential differences between SR tasks with various scale factors. Our uniform distribution training strategy can significantly improve the performance of multi-scale SR tasks. Thanks to the scale-aware strategy, our proposed continuous-scale SR model can adapt to multiple learning tasks well. Second, training with a single large-scale factor can also boost the performance on other scale factors, and the model trained on a large-scale factor has better generalization performance. This result shows that there are similarities between SR tasks with various scale factors, and it is reasonable to use a unified framework to process these multi-scale tasks.

\subsection{Feature Extractor Ablation Study} 
\begin{figure}[t]
  \centering
  \includegraphics[width=0.9\linewidth]{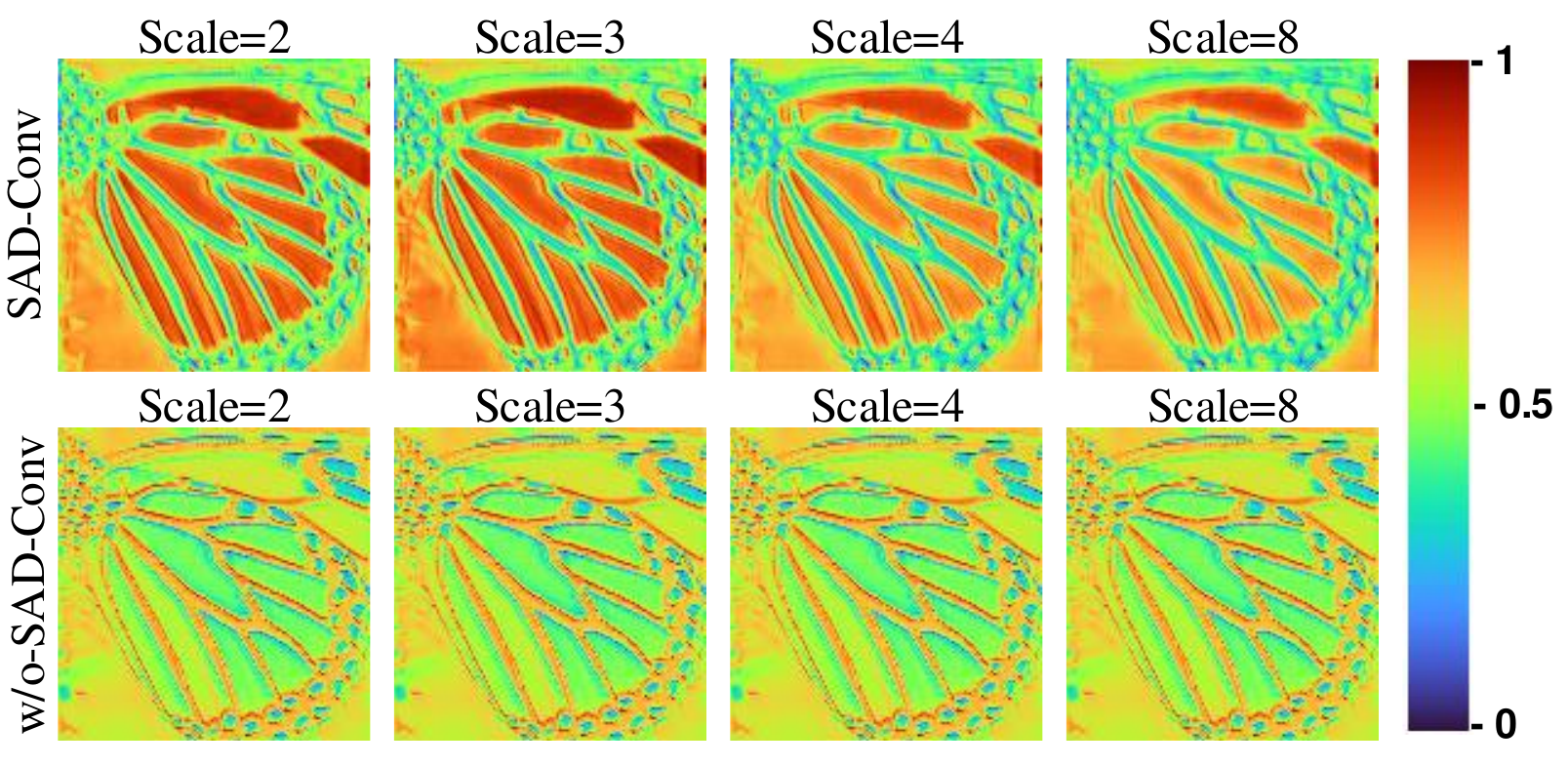}
  \caption{Average feature maps output by the SAFL modules with and without SAD-Conv. }
  \label{fig:ablation_saconv}
\end{figure}

First, to investigate the effectiveness of the scale-aware feature learning strategy, we replace SAD-Conv with a common convolutional layer with the same kernel size. Thus, a feature learning module without the scale-aware strategy is obtained. To show the difference between the SAD-Conv and the common convolutional layer, we visualize the average feature maps with different scale factors output by the SAFL modules with and without the SAD-Conv, as shown in Fig.\,\ref{fig:ablation_saconv}. We find that thanks to SAD-Conv, the SAFL can learn different feature representations based on the input scale factor while the common convolutional counterpart outputs the same feature maps for all scale factors.

Second, to explore the influence of the feedback strategy, we design a feedforward feature learning counterpart by removing all the feedback connections. Compared with the SAFL, the feedforward feature learning part only lacks a few $1\times 1$ convolutional layers for fusing the feedback features, and the number of parameters is almost the same. In this case, the feature extractor can only output one level of features as in the previous methods.

Table \ref{tab:feature_extrator_ablation} shows the PSNR comparison of different ablation models. It is observed that both the SAD-Conv and feedback strategy can improve SR performance independently. This result can be attributed to the following two aspects. First, the dynamic convolution module can provide stronger feature extraction capabilities for multi-task (multi-scale SR) learning. Second, the feedback strategy makes it possible that smaller-scale features are computed with fewer iterations while higher-scale features are refined several times from multiple recurrences. Therefore, by continuously correcting low-level features, more accurate latent codes are provided for the implicit functions, thereby improving the SR performance.

\begin{table}[tb]
  \centering
  \caption{Ablation Study on SAD-Conv and Feedback Strategy. The Models use the Same Parameters as SADN-L. The Average PSNR({\rm dB}) values are Evaluated on Set14 in $400$ Epochs.}
  \renewcommand\arraystretch{0.9}
  \begin{tabular}{|cc|rrrrr|}
    \hline
    \multicolumn{1}{|l}{SAD-Conv} & \multicolumn{1}{l|}{Feedback} & \multicolumn{1}{c}{$\times$2} & \multicolumn{1}{c}{$\times$3} & \multicolumn{1}{c}{$\times$4} & \multicolumn{1}{c}{$\times$6} & \multicolumn{1}{c|}{$\times$8} \bigstrut \\
    \hline
    \hline
    $\bm\times$                   & $\bm\times$                   & 33.29                         & 30.09                         & 28.38                         & 26.24                         & 27.74 \bigstrut[t]                       \\
    $\checkmark$                  & $\bm\times$                   & 33.38                         & 30.18                         & 28.48                         & 26.28                         & 24.78                                    \\
    $\bm\times$                   & $\checkmark$                  & 33.43                         & 30.20                         & 28.49                         & 26.26                         & 24.84                                    \\
    $\checkmark$                  & $\checkmark$                  & \textbf{33.53}                & \textbf{30.28}                & \textbf{28.55}                & \textbf{26.37}                & \textbf{24.85}\bigstrut[b]               \\
    \hline
  \end{tabular}%
  \label{tab:feature_extrator_ablation}
\end{table}

\subsection{CSUM Ablation Study}
\label{sec:CSUM_ablation}
First, we investigate the choices of local implicit feature functions in the CSUM. One of the simplest choices is to directly use the latent code with the nearest Euclidean distance as the feature of the query point. The CSUM thus defined is called CSUM-Nearest. We also use a learning-based local implicit feature function to construct the CSUM, where the local implicit feature function is a 2-layer MLP. The MLP takes the latent code set and query coordinates as input and outputs a feature vector of the query point. The CSUM thus defined is called CSUM-MLP.

Second, we investigate the scale-aware feature fusion strategy in the MBLIF. We build a CSUM without the scale-aware attention block, named CSUM-w/o-SA, in which the feature weights of all levels are the same regardless of the scale factor.

The results are shown in Table\,\ref{tab:csum_ablation}. The table shows that only using the nearest neighbor latent code to construct the MBLIF will degrade performance. Although the local implicit feature function based on the MLP seems to be more flexible, it also brings learning difficulties, resulting in a decrease in model performance. Thus, we choose the parameter-free bilinear functions to construct the MBLIF because of its high computational efficiency and effectiveness. In addition, we found that the scale-aware feature fusion strategy helps to improve SR performance.

\begin{table}[t]
  \centering
  \caption{Ablation Study on Different Choices of CSUM. The Feature Extractor of all Models is EDSR-B. The Average PSNR({\rm dB}) values are Evaluated on Set14 in $400$ Epochs.}
  \renewcommand\arraystretch{0.9}
  \begin{tabular}{|l|rrrrr|}
    \hline
    Choices of CSUM & \multicolumn{1}{c}{$\times$2} & \multicolumn{1}{c}{$\times$3} & \multicolumn{1}{c}{$\times$4} & \multicolumn{1}{c}{$\times$6} & \multicolumn{1}{c|}{$\times$8} \bigstrut \\
    \hline
    \hline
    CSUM-Nearest    & 32.88                         & 30.01                         & 27.91                         & 26.01                         & 24.09 \bigstrut[t]                       \\
    CSUM-MLP        & 33.41                         & 30.21                         & 28.48                         & 26.32                         & 24.78                                    \\
    CSUM-w/o-SA     & 33.39                         & 30.14                         & 28.47                         & 26.24                         & 24.70                                    \\
    CSUM            & \textbf{33.51}                & \textbf{30.29}                & \textbf{28.54}                & \textbf{26.38}                & \textbf{24.84} \bigstrut[b]              \\
    \hline
  \end{tabular}%
  \label{tab:csum_ablation}
\end{table}

\subsection{Number of Scales in the CSUM}

\begin{figure}[tb]
  \centering
  \includegraphics[width=0.85\linewidth]{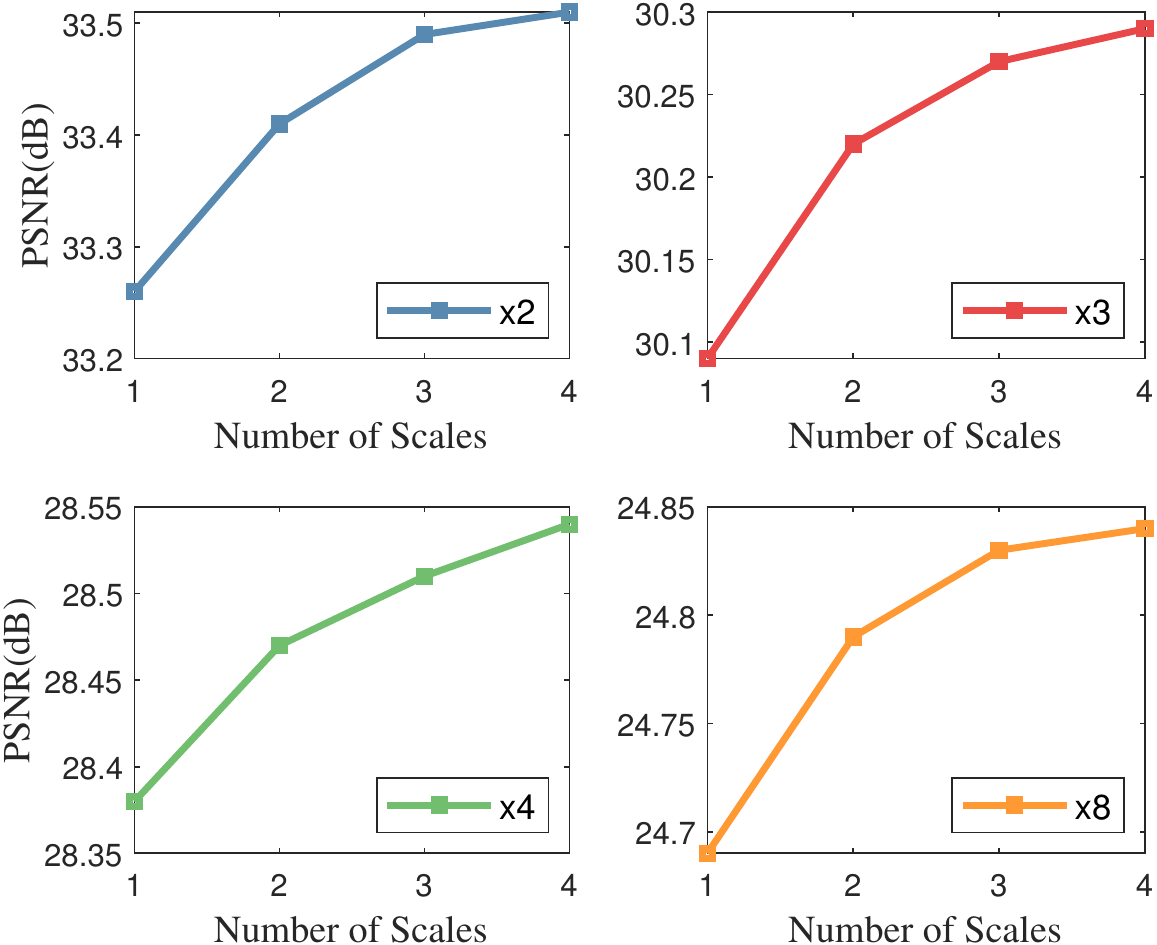}
  \caption{Investigation of the number of scales in the CSUM. Models are evaluated on Set14 over 400 epochs.}
  \label{fig:levels}
  \vspace{-1em}
\end{figure}

We investigate the number of scales $T$ in the CSUM. Because the number of iterations of the proposed SAFL needs to be the same as $T$ in the CSUM, to eliminate the effect of the feature learning part when $T$ changes, we take EDSR-B as the feature extractor in this investigation. Fig.\,\ref{fig:levels} shows that the performance continues to improve as $T$ increases from 1 to 4. This result demonstrates that using multi-scale features to construct implicit functions can significantly improve the reconstruction quality compared to using only single-scale features. In other words, our CSUM surely benefits from the multi-scale features. Although choosing a large $T$ contributes to better results, it will exponentially increase the size of the input feature maps of the MBLIF and may result in out of GPU memory. Therefore, we comprehensively weigh the performance and computational burden and set $T$ to 4 and 3 in the SADN and SADN-L, respectively.

\begin{table}[tb]
  \centering
  \renewcommand\arraystretch{0.9}
  \caption{Investigation of the number of kernels in SAD-Conv. The Average PSNR({\rm dB}) values are Evaluated on Set14 in $400$ Epochs.}
  \begin{tabular}{|l|ccccc|}
    \hline
    \multicolumn{1}{|c|}{Models} & $\times$2      & $\times$3      & $\times$4      & $\times$6      & $\times$8 \bigstrut         \\
    \hline
    \hline
    SADN-L (K=1)                 & 33.43          & 30.20          & 28.49          & 26.26          & 24.84  \bigstrut[t]         \\
    SADN-L (K=2)                 & 33.50          & 30.25          & 28.53          & 26.32          & 24.84                       \\
    SADN-L (K=3)                 & \textbf{33.53} & \textbf{30.28} & \textbf{28.55} & \textbf{26.37} & 24.85                       \\
    SADN-L (K=4)                 & 33.51          & 30.26          & 28.52          & 26.36          & \textbf{24.86} \bigstrut[b] \\
    \hline
  \end{tabular}%
  \label{tab:ablation_K}%
\end{table}

\subsection{Number of Kernels in SAD-Conv} We investigate the number of kernels $K$ in SAD-Conv. Table\,\ref{tab:ablation_K} shows the PSNRs for SAD-Conv with different $K$s. Firstly, the SAD-Conv with $K>1$ outperforms its static convolution counterpart ($K=1$) for all scale factors. Even using a small $K=2$ can significantly increase performance. This demonstrates the strength of our dynamic feature learning strategy. Second, the PSNRs stop increasing once $K$ is larger than 3. Although the representation power of the model enhances as $K$ increases, it is more difficult to learn all kernels and corresponding attention weights simultaneously, and the network is more difficult to converge.

\subsection{Investigation of Parameters Settings} 
\begin{figure}[t]
  \centering
  \includegraphics[width=0.9\linewidth]{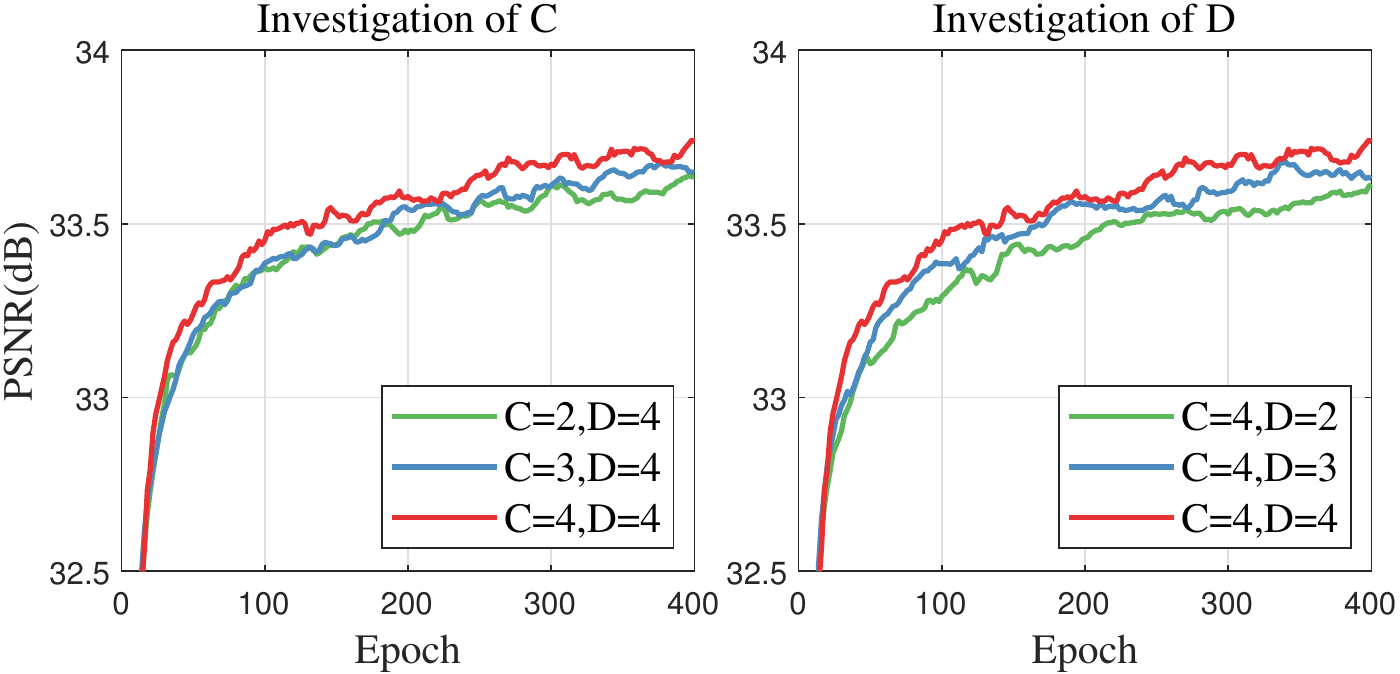}
  \caption{Investigation of $C$ and $D$. the models are evaluated on Set14 with a scale factor of $\times$2 over 400 epochs.}
  \label{fig:ablation_parameter}
  \vspace{-1em}
\end{figure}

We explore the settings of the key hyperparameters in the SAFL part, including the number of LDGs $D$ in each iteration and the number of SARBs $C$ in an LDG. We fix $C=4$ and let $D$ vary from 2 to 4. Besides, we study the influence of $C$ by fixing $D$ to 4. The results are presented in Fig.\,\ref{fig:ablation_parameter}. The figure shows that a larger $C$ and $D$ lead to better performance because a deeper network has a stronger representation ability.

\section{Conclusion}

We propose a novel scale-aware network that can dynamically perform SR with arbitrary scale factors using a single model. Compared with the previous fixed-scale SR networks, the proposed SADN can dramatically save training time and model storage space. The SAD-Conv layer in the network can adaptively adjust kernel weights based on the input scale factor, dramatically enhancing the multi-task learning ability of the network. Besides, the proposed MBLIF can make full use of multi-scale feature maps to efficiently obtain the continuous feature representation of an image, which is the core of arbitrary scale upsampling. In particular, the proposed CSUM can be easily plugged into the existing SR network to enable continuous-scale SR. The experimental results show that the proposed SADN could deliver comparative or better performance than the SOTA methods with fewer parameters and higher computational efficiency. Comprehensive ablation experiments verify the effectiveness of each component in the proposed framework.

\balance

\bibliographystyle{IEEEtran}
\bibliography{IEEEabrv,refs}

\end{document}